\documentclass[final,twocolumn, 9pt]{elsarticle}

\usepackage{lineno,hyperref}
\modulolinenumbers[5]

\journal{Journal of \LaTeX\ Templates}

\graphicspath{{./fig/}}
\DeclareGraphicsExtensions{.pdf,.png}

\usepackage{subfig}
\usepackage[cmex10]{amsmath}
\usepackage{amssymb}
\usepackage{color}

\usepackage{multirow}

\newcommand{\cb}[1]{{\cellcolor{black! #1 }$ #1$}}
\newcommand{\cw}[1]{{\cellcolor{black! #1 }$ \color{white} #1$}}
\usepackage[table]{xcolor}

\begin{document}

\begin{frontmatter}

\title{Investigating Natural Image Pleasantness Recognition using Deep Features and Eye Tracking for Loosely Controlled Human--computer Interaction}


\author[mymainaddress]{Hamed~R.~Tavakoli\corref{mycorrespondingauthor}}
\cortext[mycorrespondingauthor]{Corresponding author}
\ead{hamed.r-tavakoli@aalto.fi}

\author[mymainaddress]{Jorma Laaksonen}
\author[thirdaddress]{Esa Rahtu}

\address[mymainaddress]{Department of computer science,  Aalto University, Espoo, Finland}
\address[thirdaddress]{Center for machine vision research University of Oulu, Finland}

\begin{abstract}

This paper revisits recognition of natural image pleasantness by employing deep convolutional neural networks and affordable eye trackers. There exist several approaches to recognize image pleasantness: (1) computer vision, and (2) psychophysical signals. For natural images, computer vision approaches have not been as successful as for abstract paintings and is lagging behind the psychophysical signals like eye movements. Despite better results, the scalability of eye movements is adversely affected by the sensor cost. While the introduction of affordable sensors have helped the scalability issue by making the sensors more accessible, the application of such sensors in a loosely controlled human-computer interaction setup is not yet studied for affective image tagging. On the other hand, deep convolutional neural networks have boosted the performance of vision-based techniques significantly in recent years. To investigate the current status in regard to affective image tagging, we (1) introduce a new eye movement dataset using an affordable eye tracker, (2) study the use of deep neural networks for pleasantness recognition, (3) investigate the gap between deep features and eye movements. To meet these ends, we record eye movements in a less controlled setup, akin to daily human-computer interaction. We assess features from eye movements, visual features, and their combination. Our results show that (1) recognizing natural image pleasantness from eye movement under less restricted setup is difficult and previously used techniques are prone to fail, and (2) visual class categories are strong cues for predicting pleasantness, due to their correlation with emotions, necessitating careful study of this phenomenon. This latter finding is alerting as some deep learning approaches may fit to the class category bias.

\end{abstract}


\end{frontmatter}


\section{Introduction}
{A}{n} old desirable objective of artificial intelligence (AI) has been the understanding of emotions by machines.
Despite the exceptional recent advancements in computer science, emotion understanding still remains a challenging problem for machines. It is one of the key distinguishing factors between humans and AI. 
To reduce the gap between human and machine by incorporating emotions, the field of affective computing has been devoted to the emotion recognition and integration in human--computer interaction (HCI) scenarios. 
While the recognition of the emotional state of the users is an active area of research, the extent of emotion realization does not need to be limited to the user state. The recognition of the scene pleasantness independent of a user is an example of such a case.

Affective computing utilizes the facial expression and/or the posture of user,  physiological signals such as electroencephalography (EEG), magnetoencephalography (MEG), skin conductance response, and eye movement in order to decode the user's  state of mind~\cite{Barral2015}. On the other hand, the perception of scene pleasantness goes beyond user monitoring and can incorporate machine vision techniques to recognize the scene content in order to judge its emotional message~\cite{Yanulevskaya2008,Machajdik2010}. The first problem is well established and studied to provide user dependent emotion estimation, however, the second problem is viewer independent and is yet a challenge.

    The ability to recognize the emotional gist of a scene could facilitate more accurate emotional semantic image retrieval~\cite{Cho2004,Wang2008}. Further potential applications can include image/video content analysis and annotation, affective assessment and refinement of advertisements, multimedia affective rating, etc. Motivated enough by such applications, in this work, we focus on decoding the pleasantness of  images. The image pleasantness or {\it valence} is the indicator of the amount of positive expression or the negative effect of an image. In order to estimate the pleasantness, we employ computer-vision-based techniques and user-in-the-loop methods as well as the combination of the two procedures. 

\subsection*{Contributions:}

Our main contribution is the assessment of image pleasantness recognition in a realistic, less constrained human-computer interaction scenario, in which the most recent computer vision state-of-the-art and eye tracking based techniques are challenged. More precisely, our contribution includes:

\begin{itemize}
\item We provide a new data set consisting of eye movements of 20 observers for 382 emotional images. Contrary to the existing data sets, which are often relying on high-end expensive devices to record the eye movements, we utilize an affordable inexpensive device. This helps us to be as close to an everyday HCI scenario as possible.
\item It is often argued that the lack of mid- and high-level features has an adverse effect on successful scene pleasantness decoding using conventional features for machine vision methods. The deep convolutional neural networks (CNNs), however, address this issue. We assess the performance of the deep CNNs for image pleasantness recognition and provide a comparison with the existing vision-based techniques.
\item We investigate the replicability of valence decoding from eye movements using affordable eye tracker and a less constrained setup. 
To this end, we use the 95 images, employed by our former study~\cite{RezazadeganTavakoli2014}, with new eye movement data.
\end{itemize}

In the rest of this paper, we first overview the related work. Then we introduce a new dataset including stimuli, setup and procedure for recording eye movements, and analyze the statistics of the collected data. In Section~\ref{sec:ML}, we elaborate the basis of a machine learning approach in which classifiers are trained with various visual and gaze-based features in order to decode the pleasantness category of an image. Section~\ref{sec:EXP} contains several experiments to investigate the performance of features using the adopted machine learning approach. Afterwards, we discuss the results, followed by conclusions.

\section{Related Work}

\subsection{Psychology} 
Psychology has been the driving force of emotion understanding studies and the inspiration of existing HCI methods. The theories on the emotional message of colors~\cite{Itten1973} are among influential studies that  inspired the early works in image pleasantness recognition~\cite{Colombo1999}. 
The color theory associates red with positive emotional impact, whereas purple is related with negative emotional feelings. The color theory has evolved through the years and emerged in various psychological models of color emotion~\cite{Ou2004}.
The color theory has been applied to the study of natural and abstract scenes, albeit the study of abstract images mostly benefits from it.

In the span of natural images, the \emph{International Affective Picture System} (IAPS)~\cite{Lang2008} and its related research studies are among the most well-recognized ones in the computer science community. The span of such studies covers a wide range, including the correlation of motivation and attention with emotions~\cite{Lang1995}, the affect of scene complexity on emotional arousal~\cite{Bradley2007}, and sex differences in emotion perception~\cite{Bradley2001}.

The psychology have investigated various properties of images and their emotional affect on observers under different theories such as ~\emph{gestalt}~\cite{Arnheim1974}. There exists studies~\cite{Bar2006} that investigate the effect of curves on human perception, which are considered appealing. Conversely, chaotic texture or angular and diagonal patterns are considered to evoke negative emotions.

Psychological studies also include the study of different physiological signals, e.g., eye movements. These studies mostly demonstrate the influence of emotions on eye movement patterns. The comparison of observers' eye movements on emotionally different images reveals distortion in the patterns during exposure to highly emotive images~\cite{Nummenmaa2006}. The assessment of the performance of emotive stimuli detection by observers is another indicator of emotion influence on the eye movements, which is highly studied in psychology~\cite{Humphrey2012,Niu2012a}.

\subsection{Visual Features}

The early works, based on visual features, are mostly inspired by the color theory. In a semantic information retrieval task, a series of color-based features are used in~\cite{Colombo1999}. The authors decomposed an image into homogeneous regions and computed the hue, brightness, and saturation as well as the position and the size of each region. Then using a framework of compositional semantics, they performed a bottom-up analysis in a two-level hierarchy to combine the visual features into an expressive feature set, which is hypothesized to be equivalent with the high-level information abstraction in the human brain. Eventually, they employed such features in associating images with semantical terms in order to retrieve images with specific emotional gist.

In ~\cite{Wang2005}, an emotional image retrieval system was developed using color semantics. They built a dictionary of color to emotion mappings by clustering the color descriptors using a fuzzy clustering scheme and implementing a fuzzy system in order to relate colors to emotions. The association of each image segment to an emotion in the dictionary is then decided by the amount of the membership of each pixel, calculated using the fuzzy system. This produces a regional semantic descriptor which is augmented by a descriptor based on the average lightness, average saturation, and the average color contrast of the whole image in order to perform emotional semantic queries.

Later, the color emotion was used for both image emotion classification and retrieval in~\cite{Solli2008}. In fact, they extended the existing methods by enabling submission of a query image instead of emotion scales. To this end, a kd-tree decomposition over a color-based emotion space was used to map a given image to some emotional category. The same idea was later developed using a bag-of-emotions approach~\cite{Solli2009}.  The bag-of-emotions is a histogram of the number of occurrences of particular emotion patterns. 

While the aforementioned techniques rely on some psychological prior assumptions~\cite{Ou2004} to define a color-based emotion space, later studies  started learning the association of color and evoked emotions directly from labeled images. For example,~\cite{Yanulevskaya2012} applied a bag-of-visual-colors to learn the emotional gist of abstract paintings from the emotional ratings of users. A similar concept was utilized by~\cite{Sartori2015} using a sparse lasso regression technique. We are seeking a similar approach in employing visual features and extending them to include CNNs.

The visual features are not limited to color and can include texture and shape. Thus, it is possible to utilize, for example, a series of holistic features based on Gabor filters~\cite{Bovik1990} and Wiccest features~\cite{Vailaya2001}, that encode the edginess and texture characteristics, in order to infer the emotional category of images~\cite{Yanulevskaya2008}. In~\cite{Lu2012}, it is tried to capture the scene pleasantness from shapes by using shape-based features in emotion category decoding. They exploited features extracted from line segments, angles, continuous lines, and curves.

The emotion elicitation, however, is not limited to the effect of low-level features discussed above. 
The higher-level semantic content within an image, such as faces, text, animals, objects, the amount of skin and so on, and their interaction with themselves and the environment, is by far recognized as a crucial missing component~\cite{Lew2006}.
Particularly, in the case of natural images, content-related features matter. Motivated by such observations,~\cite{Machajdik2010} studied the role of content-related features. They revealed that various content-related features can contribute to the emotional gist of the scene. Eventually, they argued that semantic content analysis is crucial in the recognition of the affective categories of images.

Recently, deep Convolutional Neural Networks (CNNs) have influenced the  computer vision research substantially. There exist numerous successful applications, such as image and scene classification~\cite{Krizhevsky2012,Koskela2014}, object detection~\cite{Szegedy2013}, and object segmentation~\cite{Long2015}, where the state-of-the-art relies on deep CNNs.  During the preparation of the final manuscript, we learned that~\cite{You2015} and~\cite{Wang2016} used CNN features for pleasantness recognition, under the umbrella of sentiment analysis, to infer if an image is pleasant or unpleasant. In other words, they are solving a two-class, positive and negative, classification problem. 
While~\cite{You2015,Wang2016} apply an end-to-end approach, that is, they learn the CNN features and task together, we utilize a filter bank approach as in~\cite{Cimpoi2015} because of the relatively small number of images in our data set. There are also other differences in our problem setup: 1) we use a three class classification, \emph{unpleasant}, \emph{neutral}, and \emph{pleasant} paradigm, which in our opinion is more natural, and 2)  in our data set, the ground truth emotional labels are based on the diverse scores of \emph{International Affective Picture System} (IAPS) corpus~\cite{Lang2008} ratings, while the data used by\cite{You2015,Wang2016} uses a majority vote scheme of at most five people to decide the emotional class categories.

\subsection{Eye Movements}

The automatic detection of the semantic content is often difficult, and sometimes replaced by other means such as observers' eye movements. For example,~\cite{Subramanian2011} applied eye movements as a proxy to identify scene content such as interacting elements. Similarly, a method for localization of affective objects using eye movements was developed in~\cite{Subramanian2009}.

The eye movements, however, are not only a cue for semantic content decoding. There exist studies which address the user independent recognition of the pleasantness of images and videos using various physiological signals including eye movements. In these approaches, the eye movement is a signal, gathered unobtrusively from several observers and aggregated in order to tag an image implicitly and independent of a specific user. For example,~\cite{Soleymani2012} utilized EEG and eye tracking for affective video tagging in which they relied on features extracted from pupil diameter, gaze distance, and eye blinking obtained from gaze recordings. 

In~\cite{RezazadeganTavakoli2014}, eye movements were applied  in order to recognize the emotional message of an image in terms of its pleasantness. To this end, several features were utilized including fixation duration, fixation location, saccade slope, saccade length, saccade orientation, and saccade veloctiy in order to build a classifier. Eventually, it has been demonstrated that eye movements outperform the bag-of-visual-colors~\cite{Yanulevskaya2012} and visual SIFT~\cite{Lowe1999} features in the case of abstract images. 

Later, the influence of the features extracted from eye movements and their contribution for image pleasantness recognition were studied in~\cite{R.-Tavakoli2015}. Utilizing various feature selection techniques, it was demonstrated that the most influential features for such a task are the fixation duration and fixation density map. Furthermore, the analysis of feature representation schemes confirmed that the histogram-based feature representation have the edge over traditional average-value representations.

In regard to the eye movement features, we are extending the works of~\cite{RezazadeganTavakoli2014} by incorporating more images with eye movements. There exists a major difference that is on the sensor side and data. Instead of a high-end accurate eye tracker that is not available to everyone, we are utilizing an affordable sensor available to any user. We, further, adopt a less restrictive experiment setup to mimic user interaction in an everyday computer use setting where there is no expert guiding the calibration and users perform the calibration themselves upon training. 

\section{Data Set}

The data set consists of affective images and corresponding eye movements of several observers for each image. The eye movements are recorded in a free-viewing task. We first explain the stimuli and its characteristic.
Then we elaborate the experiment setup, procedure, and task. Finally, we analyze the recorded eye movements using conventional statistical analysis approaches.

\subsection{Stimuli}
We have a total of 382 affective images, all selected from the \emph{International Affective Picture System} (IAPS) corpus~\cite{Lang2008}\footnote{The images can not be shown due to copyright restrictions imposed by the image corpus. Please visit~\url{http://csea.phhp.ufl.edu/media.html} to obtain the images from the center for the study of emotion and attention, University of Florida.}.  
The images have the resolution of $1024\times768$. The image set includes the same 95 images which were previously used in the studies of emotional valence recognition~\cite{RezazadeganTavakoli2014}. There is gender-specific emotional rating agreement and only one class of visual content, labeld by human, for the 95 images. The set of 382 images, however, includes a wide span of emotional and visual contents including reptilians, wild predators, domestic animals and pets, people and activities, portraits, erotic and nude, objects (e.g., cup, stool, etc), foods, events (e.g., flood, explosion, protest, etc). 

The 382 images are accompanied with the emotional valance ratings in the form of mean self-assessment manikins (SAM)~\cite{Bradley1994} score per image in which the score of 1 indicates the most unpleasant case and 9 is the most pleasant image. The mean SAM score of each images is provided in terms of the genders and all users. Assigning images with mean valence value in the range of $4-6$ as neutral, the image set consists of 134 pleasant ($\mu=6.96,\sigma=0.61$), 84 unpleasant ($\mu=2.94,\sigma=0.64$) and 164 neutral images ($\mu=5.10,\sigma=0.55$). 
Based on the gender-specific emotional ratings of the IAPS, there is a strong agreement between genders on the emotional content of only  296 of the images; from which 102 are pleasant ($\mu=7.16,\sigma=0.53$), 68 are unpleasant ($\mu=2.78,\sigma=0.56$) and 126 are neutral ($\mu=5.04,\sigma=0.47$).

\subsection{Observers}
The participants are 20 volunteers, 12 male and 8 female, with mean observer age of 29.9 (std=7.36, min=20, med=27, max=53). They are graduate and postgraduate students majoring in computer science. The participants have normal or corrected to normal vision and never reported having eye-sight problems, nor any psychological disorders. They have not previously seen the stimuli. Among the images 15 are always displayed twice, i.e., the participants watch 397 images, though the second run of an image is not used in our experiments. Each participant views all the images in one session.  The whole session including the instructions does not exceed 1 hour.

\subsection{Eye Tracking Procedure}

We are using a Tobii EyeX eye tracker, which is one of the most affordable devices (less than 100 Euro in 2014), to record the eye movements. 
The information regarding the configuration and specification of the device is summarized in Table~\ref{table:eyedevice}. We use the provided API in order to determine the gaze location and fixation events using the default recommended parameters.
 The images are displayed on a 19 inch LCD at the  resolution of $1600 \times 1200$. The images are screened at the center of a black background in their original resolution ($1024\times768$). Each image is presented for 5 seconds followed by a gray mask for 2 seconds. To be less restrictive, the observers are distanced at about 65--70 cm from the monitor at their convenient sitting posture and no chin rest is used. They are instructed to keep their heads within the tracking plane, which provides the freedom of small head movements, and to watch the images during the viewing time. The calibration procedure is 9-point.

Since we would like to be as close to an uncontrolled environment and mimic unsupervised interactions as in daily life,
each observer is initially given instructions on using the eye tracker and helped with the calibration procedure. Once the users know how to use the device, they are left to repeat the calibration procedure once again themselves, as expected in a real-world application scenario, and run the viewer application. The viewer application simply presents the images as described above and records the users' fixation and gaze information.

\begin{table}[!t]
\tiny
\renewcommand{\arraystretch}{1.3}
\caption{Tobii EyeX Tracking Device Specification and Configuration.}
\label{table:eyedevice}
\centering
\begin{tabular}{|l||r|}
\hline
{\bf Specification \& Configuration} & {\bf value} \\
\hline
\hline
Sampling rate & $>60$Hz \\
Latency & 15 ms +/- 5 ms \\
Operating distance & 45--100~cm\\
Headbox size & $40 \times 30$~cm at 65~cm\\
Firmware & 2.0.2-33638\\
Core version & 2.0.9 \\
Driver version & 2.0.9 \\
Service version & 1.9.4.6493 \\
Engine version & 1.9.4.6493 \\
Configuration version & 3.2.9.521 \\
Interaction version & 2.1.1.3125\\
C++ SDK version & 1.7\\
\hline
\end{tabular}
\end{table}

\subsection{Statistics}

\begin{figure*}[!t]
\tiny
\centering
\subfloat[]{
\includegraphics[scale=0.25]{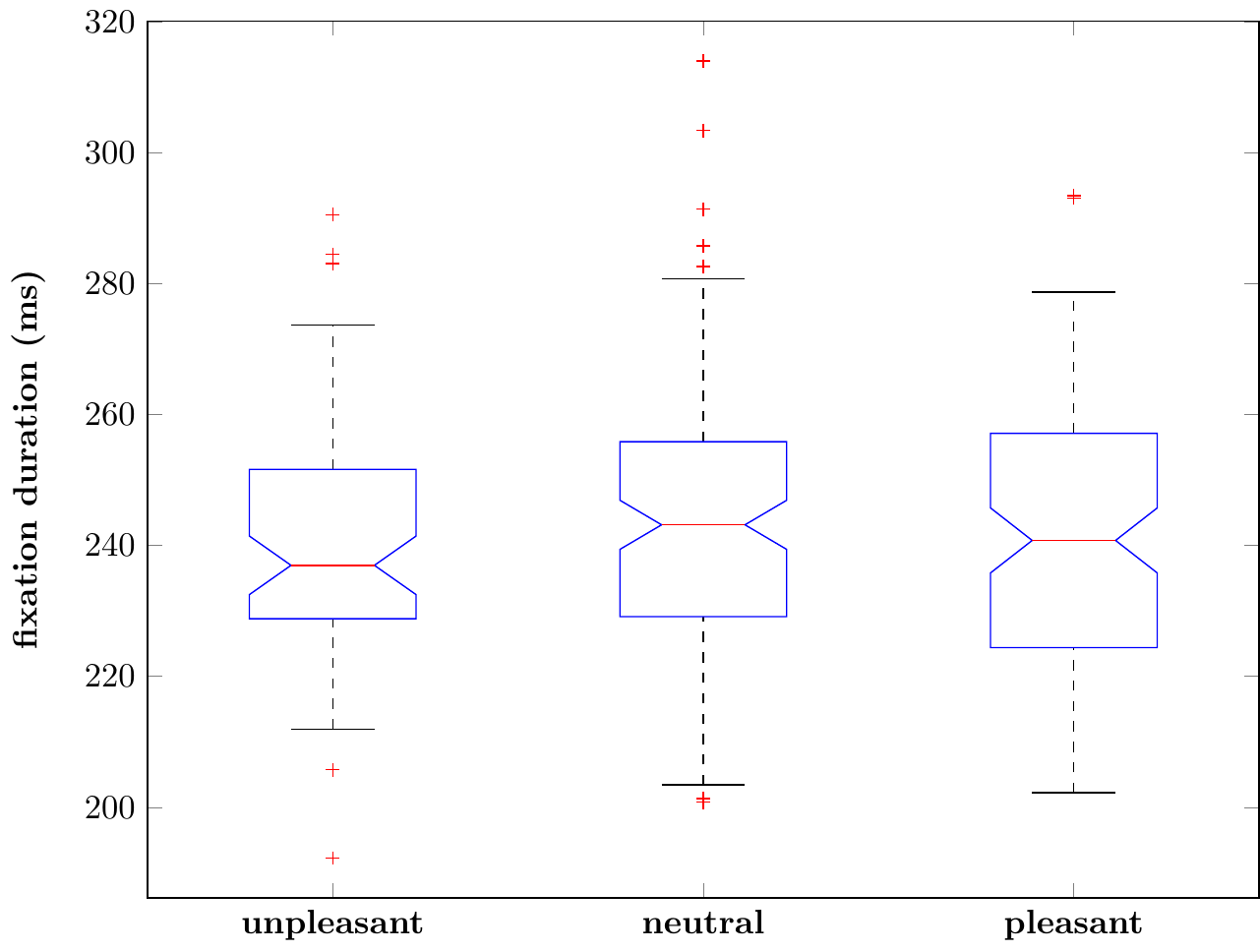}
\includegraphics[scale=0.25]{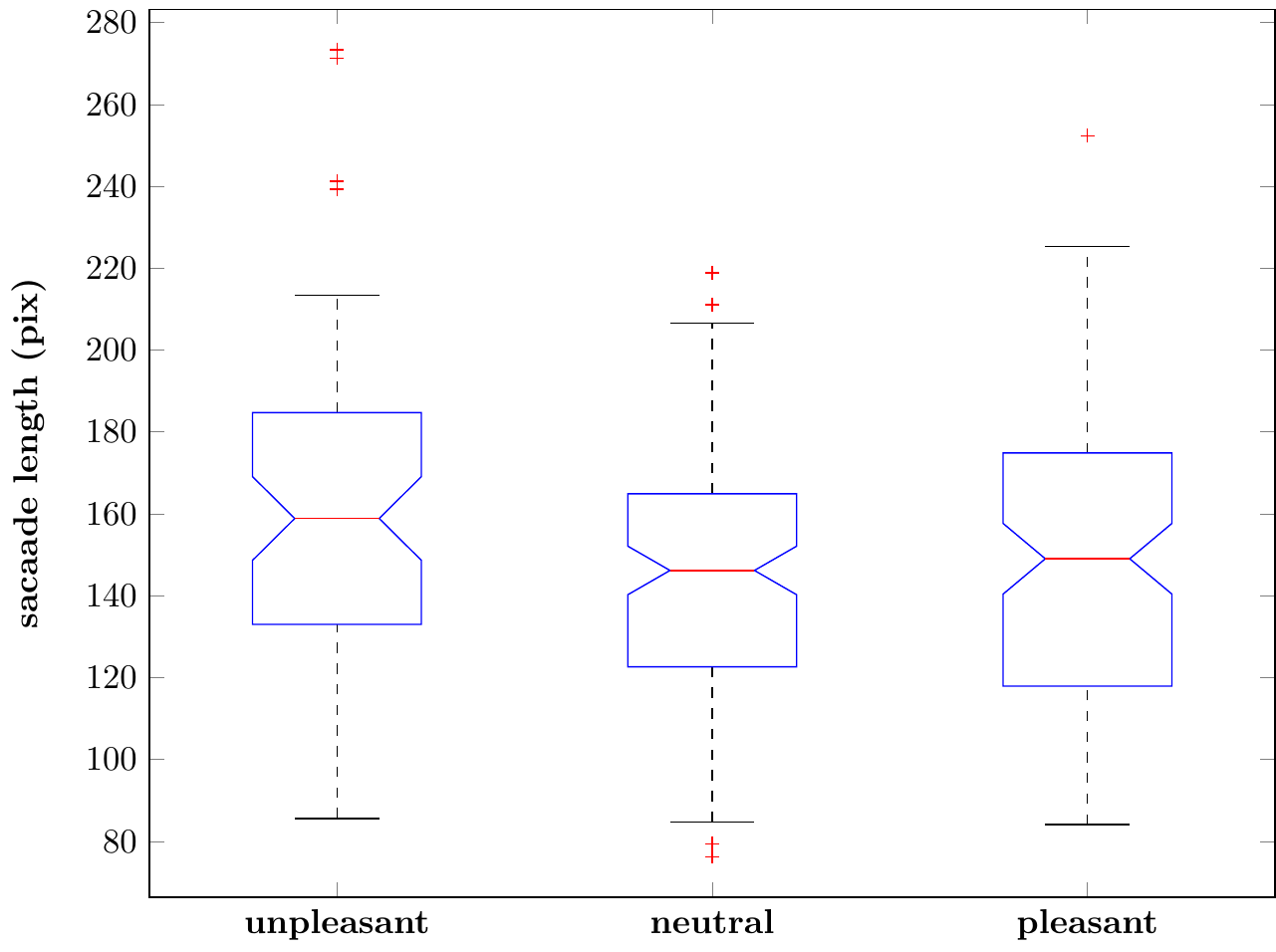}
\includegraphics[scale=0.25]{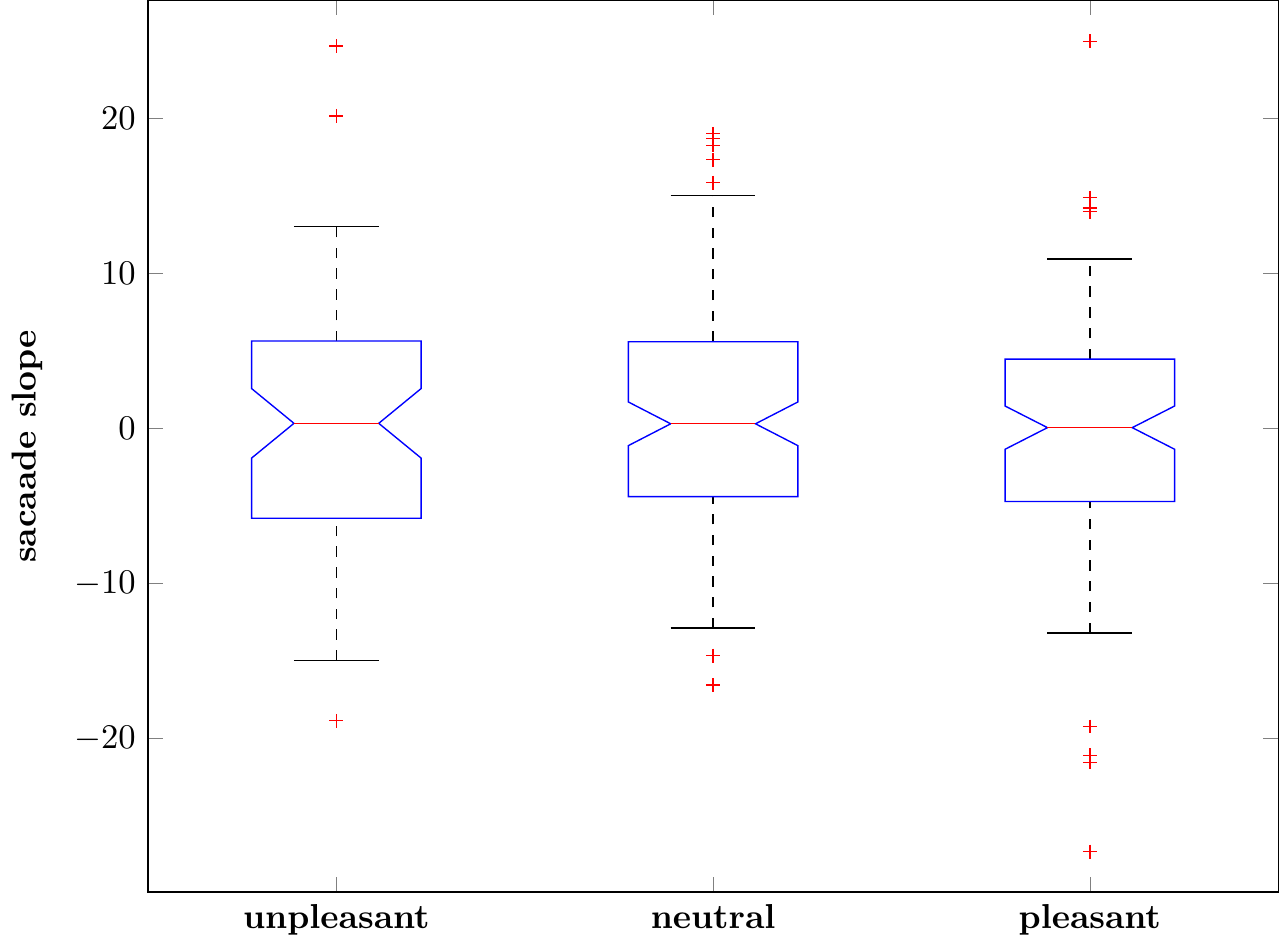}
}
\subfloat[]{
\includegraphics[height=1.0in, width=1.5in]{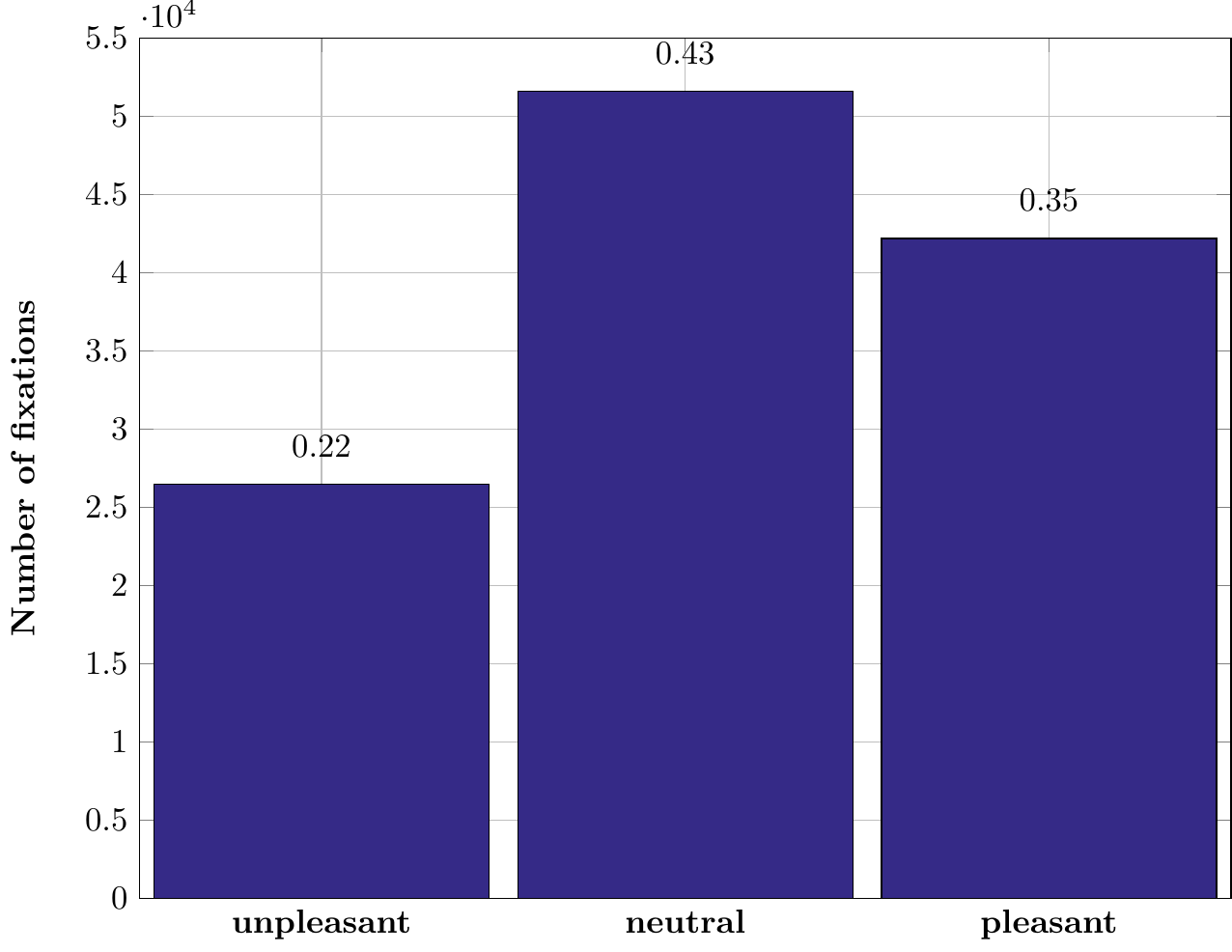}
}
\caption{Eye movement statistics: a) box plots for fixation duration, saccade length, and slope, for different emotional categories, b) Number of fixations per emotional category with their ratio to the total number of fixations.}
\label{fig:fixBoxPlot}
\end{figure*}

The data set includes approximately 7 hours and 44 minutes of gaze data consisting of 120219 fixations. 51574 of fixations are on neutral images, 42173 on pleasant, and 26472 on unpleasant images. To gain further insight about the data, we looked into some influential characteristics, which are repeatedly reported of having a role in identifying the affective state of observers and image pleasantness tagging, namely ``fixation duration'', ``saccade slope'', and ``saccade length''~\cite{Wadlinger2006,Soleymani2012,Tichon2014,Simola2015,R.-Tavakoli2015}. For this purpose, we pulled the data of all the observers together and computed the average feature per image in each emotion category.
Figure~\ref{fig:fixBoxPlot} presents the box plots of the fixation duration, saccade length, and saccade slope for the unpleasant, neutral, and pleasant  images along with the number of fixations in each emotion category. Using this data, we did an ANOVA test,  which indicates that there is no significant difference ($p > 0.05$) between the three emotional classes in terms of average fixation duration for $[ F(2,379)=0.33, p=0.72]$ and average saccade slope for $[ F(2,378)=0.13, p=0.88]$. The average saccade length is, however, significantly different ($p < 0.05$) for emotion categories for $[ F(2,378)=4.42, p=0.012]$. The Tukey-Kramer's post-hoc test reveals that the saccade length significantly differs for unpleasant images compared with neutral ones, while there is no significant difference between the neutral and pleasant images as well as for the pair of pleasant and unpleasant images.

We also repeated the above analysis after removing the images with strong disagreement between the genders. The results are similar, i.e., for average  fixation duration there is still no significant difference ($p > 0.05$) between the three emotional categories for $[ F(2,293)=0.49, p=0.61]$ and also for saccade slope $p > 0.05$ for $[ F(2,292)=0.56, p=0.57]$. The mean average saccade length is significantly different ($p < 0.05$) for emotion categories for $[ F(2,293)=4.17, p=0.02]$, where the Tukey-Kramer's post-hoc test again shows the saccade length significantly differs between unpleasant and neutral images with no difference between the pair of neutral and pleasant images as well as the pair of unpleasant images and pleasant ones.

To examine the possible differences between genders~\cite{Bradley2001,Lithari2010} in the current data, we pulled the data of each gender category and compared them against one another using ANOVA. The results indicate that for the current data, there is no significant difference ($p > 0.05$) between the genders in terms of average fixation duration for $[F(588,1)=2.27, p=0.13]$ and saccade slope $[F(587,1)=0.15, p=0.69]$. There is, however, a significant difference ($p < 0.05$) between the genders in terms of mean saccade length $[F(587,1)=57,72, p\approx0.0001]$, i.e., the saccade length statistic is different between male and female participants per image.

An interesting statistic to check is the variability among subjects in looking at the same images. To measure such a variability, we compute the inter-observer visual congruency~\cite{Torralba2006} for the data. For this purpose, we use fixation locations and adopt a leave-one-out policy as in~\cite{Torralba2006}. We leave out one participant and build a fixation density map from the rest of the fixations. Then, we assess how well the fixations of the left-out participant consists with the rest of the observers using the area under the curve (AUC) metric. For an image, the average of the AUC score of the participants is recognized as the inter-observer visual congruency (IOVC) score of that image. We compute the mean IOVC score for all the images and each image category. The IOVC for all the images is 95.34\% +/- 1.9. Looking finer into the images in each emotion class category, the IOVC values are 95.45\% +/- 1.5 for the unpleasant, 95.43\% +/- 1.7 for the neutral, and  95.16\% +/- 1.8 for the pleasant images. Apparently, there is no significant difference ($p > 0.05$) between emotional class categories in terms of mean IOVC of images for [$F(2, 379)=1.15, p=0.32$].

\begin{figure}[!t]
\centering
\subfloat[]{
\hspace{1.5mm}
\includegraphics[height=0.9in, width=1.2in]{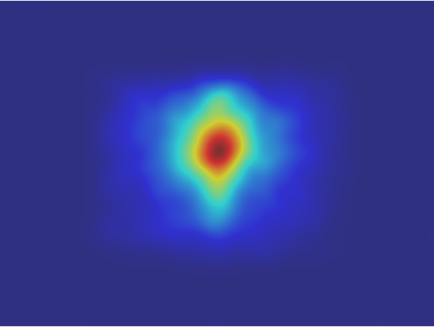}
}
\subfloat[]{
\includegraphics[height=0.98in, width=1.2in]{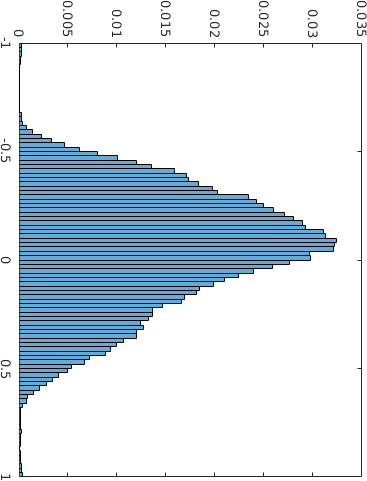}
}

\subfloat[]{
\includegraphics[height=0.9in, width=1.3in]{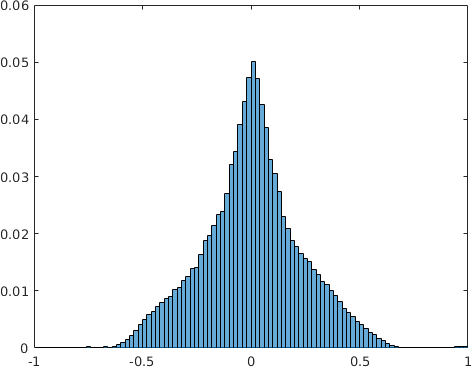}
}
\subfloat{
\includegraphics[height=0.9in, width=1.2in]{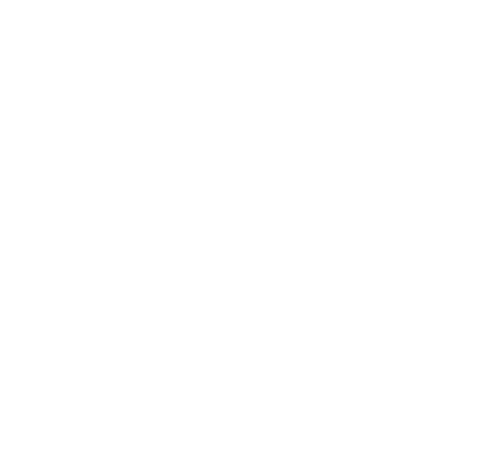}

}

\caption{Center bias. a) The mean eye position map (MEP), obtained from smoothing all the fixations of all the participants on all the images. b) The vertical distribution of the fixations. c) The horizontal distribution of the fixations. }
\label{fig:centerBias}
\vskip -6pt
\end{figure}

We also measure the amount of center bias~\cite{Tatler2007} in the database and in each emotional class category. We gathered all the fixations together to obtain a fixation map. Afterwards, a two-dimensional normal distribution $\mathcal{N}(\mu, \Sigma)$ is fitted, where $\mu=[\mu_1, \mu_2]$ is the average location and $\Sigma = \mathbf{diag}(\sigma_1^2, \sigma_2^2)$ is the covariance of the locations. Figure~\ref{fig:centerBias} depicts the center bias for the data set in terms of mean eye positions, the horizontal and vertical projections of the probability distribution of fixations within the normalized image size of [-1,~1], where (0,~0) is the image center. The parameters of the estimated distribution are $\mu = [0.006, -0.046]$ and $\sigma = [0.23, 0.26]$. Therefore, the fixations are a bit deviated upwards\footnote{The system coordinates used are compatible with left-handed coordinates as in Computer Graphics.}. It is not surprising that similar to the previous research in visual perception, the current data is also affected by the center bias phenomenon.

\section{Method}
\label{sec:ML}

Nowadays, the use of machine learning approaches in analyzing physiological signals have become a standard norm and most recent studies of eye movements include some machine learning technique to prove a hypothesis, e.g.,~\cite{Greene2012,Borji2014a,Kanan2014,Simola2015}. This approach has been also favored in the HCI community for a long time and provides the flexibility of integrating various signals and features into a system. In this section, we first explain the feature extraction, which includes both features extracted from eye movements and images, followed by the classification scheme.

\subsection{Eye Movement Features}

It has already been demonstrated that various eye movement features contribute to the recognition of the emotional message of a scene. 
We employ a selection of features based on aforementioned fixation and saccade characteristics, including fixation locations, fixation duration, saccade length, and saccade slope.

The fixation location is encoded in terms of fixation density maps. The fixation density map feature is computed by fitting a mixture of Gaussian kernels at the fixation locations. Then the map is down-sampled to the size of $20\times15$ and vectorized to build a feature vector.
To avoid scene encoding, which the fixation density map is susceptible to be affected by, akin to~\cite{R.-Tavakoli2015}, we also study the entropy of the fixation density map.
The entropy of the fixation density map indicates the amount of information a map carries independent of the exact locations of the fixations.

For the rest of the gaze properties, we use the mean value and standard deviation as the feature representation similar to~\cite{Greene2012}, as well as the histogram representation of~\cite{R.-Tavakoli2015}. The histogram of fixation duration is computed by quantizing the possible range between the minimum and maximum duration into 60 bins. The number of bins is chosen in relation to time quantiles. Then the number of the occurrence of each bin is counted. A similar approach is adopted to make a histogram of saccade slope and saccade length. The saccade slope histogram consists of 30 bins with quantiles of $6^{\circ}$ for the values in the range of $0^{\circ}$ to $180^{\circ}$. The histogram of saccade length consists of 50 bins.

As an alternative to the above feature representations, which explicitly exploit the properties of eye movements, we also employ an automated feature learning technique based on Fisher Kernel Learning (FKL)~\cite{Maaten2011}. The FKL algorithm treats the eye movements as a time-series and tries to automatically learn a hidden Markov model representation that maximizes the Fisher criterion. This approach is successfully used for observer task detection from eye movements by~\cite{Kanan2014}. Similar to~\cite{Kanan2014}, we employ 10 hidden states and center the data with respect to the training set.

Another fixation-based feature is inter-observer visual congruency. Its emphasis is on the differences between the observers on watching an image.
The inter-observer visual congruency have been applied for image ranking~\cite{LeMeur2011a} and image memorability prediction~\cite{Mancas2013}. We, here, adopt the concept to build a feature vector consisting of the observer visual congruency scores for images. For each image, we compute the consistency of the observers' eye movements in terms of AUC and report the mean and standard deviation as features to build a 2-dimensional feature vector. It is worth-noting that the proposed feature vector can be seen on as a compressed version of the fixation density map vector which signifies the degree of observer agreement on an image.

\subsection{Visual Features}

The role of visual elements of a scene can not be neglected. 
It is often speculated that the computational visual features often fail because they can not fully model the high-level semantical information encoded in an image~\cite{Lew2006}.
Nonetheless, recent advancements in the area of deep neural networks has introduced some hope and expectations for success in the task of image pleasantness recognition.
Unlike the engineered features such as SIFT~\cite{Lowe2004}, PHOW~\cite{Bosch2007}, and SURF~\cite{Bay2008}, a deep network encodes low-, mid-, and some degree of high-level contextual information. 
In our system, we use visual features including PHOW features and deep features to assess the performance of the deep features in comparison to other visual features and eye movements.
 
\subsubsection*{PHOW Features}
 
 Motivated by the characteristic of the  SIFT in capturing contours and edges,\cite{Yanulevskaya2012} and~\cite{RezazadeganTavakoli2014} applied SIFT descriptors for analyzing the emotional gist of an image. They applied densely extracted RGB-SIFT and RGI-SIFT~\cite{Sande2010} in order to obtain an image representation.
 
To incorporate multiple scales, we adopt the Pyramid Histogram of visual Words (PHOW) descriptor~\cite{Bosch2007}.
It is  a variant of the dense SIFT descriptor which extracts SIFT descriptors at several scales and can be easily adopted to support color features. In particular, we apply the PHOW-color to incorporate color information descriptors.
Finally, we encode each image using a Bag-of-Visual-Words (BOW) consisting of a
visual vocabulary of 4096 words obtained using a k-means clustering scheme. The vocabulary size is selected not to be too small, nor too large to prevent bad representativeness and overfitting, respectively. The number of 4096 words is the typical vocabulary size for successful and efficient image recognition tasks.
It is worth noting that the PHOW-color variant almost corresponds to the same features used in~\cite{Yanulevskaya2012,RezazadeganTavakoli2014}, and~\cite{Sartori2015} and can be considered a replicate of their visual features.
 
\subsubsection*{Deep Features}

Despite the training of CNNs often requires gigantic amount of data, it is possible to apply the features extracted using pre-trained neural networks for a small amount of test data due to the generalization properties of the deep feature extractors~\cite{Cimpoi2015}. To perform such a domain transfer, the activations on the output of a deep, fully-connected layer is often combined with a target domain specific classifier. The number of images in our data set is relatively small compared to the number of images needed for training a deep CNN or even fine-tuning a pre-trained network for a specific task. Thus, we adopt a bank of filters approach to CNNs~\cite{Cimpoi2015}. For this purpose, we employ the very deep convolution networks~\cite{Simonyan2015}, recognized as the VGG architecture, which is shown to be useful in filter bank approaches~\cite{Cimpoi2015}.

The input to the architecture is a fixed-size, mean subtracted image. The image goes through the pre-trained network which consists of convolutional layers with $3\times3$ receptive fields and max-pooling with the stride of 2 pixels between layers. The convolutional layers are followed by two fully-connected layers each having 4096 outputs. We use the output of the last layer as the features in our framework\footnote{Note: the architecture has 3 fully-connected layers, where the last layer performs classification and the other layers can be considered as transformations from the convolution pipeline to a compact feature vector representation. We have reported only the layers utilized.}, which is equivalent to a compact feature representation.
The performance of the architectures with 16 (VGG16) and 19 (VGG19) layers are studied. 
 
\subsection{Classification Scheme}
 
The classification is carried out using support vector machines (SVM)~\cite{Cortes1995}. Following~\cite{R.-Tavakoli2015}, which compared the linear, polynomial, and radial basis SVMs and concluded linear SVM is a proper choice for emotion recognition, we employ linear SVM as the classifier. The classification scheme is one-versus-rest where there exists one classifier for each image pleasantness category. To evaluate the performance, we perform a repeated 10 fold cross-validation, with three sets of \emph{train}, \emph{validation}, and \emph{test} of the ratios of 0.9, 0.05, 0.05, respectively. To deal with the imbalanced data, in each repetition and fold, we guarantee an equal number of samples from each class category following a resampling with partitioning strategy~\cite{He2009} as in~\cite{Atyabi2012}. 
Consequently, the chance level for a three class classification is thus 33.3\%.
The results are summarized as confusion matrices and mean accuracies (mA). The 95\% confidence interval (CI) is also reported. We further perform a McNemar test~\cite{McNemar1947} to evaluate how different the performance of classifiers is from chance.

\section{Experiments}
\label{sec:EXP}

To conduct the experiments, we consider three scenarios for the stimuli/images. These scenarios are 1) 95 images which have high emotional agreement between genders and narrowing down the image visual class categories to one class category as in~\cite{RezazadeganTavakoli2014}, that is the visual class of ``People \& Daily Activity'', annotated by human, 2) 296 images, which are emotionally agreed between the genders, but include any visual class category, 3) the whole 382 image set, where some of the images are not emotionally agreed between genders and a diverse set of visual class categories also exist. Depending on the goal of the experiment, we will use one or all the three scenarios.

\subsection{Eye Movements}

We first perform the experiments with the subset of 95 images used in~\cite{RezazadeganTavakoli2014} and learn a classifier to categorize them into three emotional classes of pleasant, neutral, and unpleasant. Figure~\ref{fig:confusion00} summarizes the results, showing that none of the features based on eye movements perform significantly above chance level (33\%). Then, we extend the image set to the 296 images which are agreed between the genders and learn a classifier to categorize the images into the three possible classes. For this purpose, we use each of the proposed features individually. The result is summarized in Figure~\ref{fig:confusion0}. As depicted, it is only the fixation density map that performs significantly better than chance at 37.8\% [$p=0.01$, 95\% CI=35.65--39.45], with a well-balanced confusion matrix where all the emotional class categories are distinguishable above chance.

\begin{table*}
\tiny
\centering
\begin{tiny}
\captionsetup[subfigure]{labelformat=empty}
\subfloat[\tiny mean+std fixation duration]{
	\begin{tabular}{c | l c c c }
		\multicolumn{1}{c}{} & & \multicolumn{3}{c}{Prediction} \\ \cline{3-5}
		\multicolumn{1}{c}{} & & unpleasant & neutral & pleasant \\
		\multirow{3}{*}{\rotatebox[origin=c]{90}{Actual}}		
		& unpleasant	& \cw{32.1} & \cb{33.6} & \cb{34.3} \\
		& neutral	& \cb{30.1} & \cw{35.0} & \cb{35.0} \\ 
		& pleasant	& \cb{36.6} & \cb{31.9} & \cw{31.5} \\
		\multicolumn{5}{r}{}\\
		\multicolumn{2}{r}{} & \multicolumn{3}{c}{mA=32.9\%}\\
		\multicolumn{2}{r}{} & \multicolumn{3}{c}{($p=0.86$, 95\% CI=29.55--36.11)}
	\end{tabular}
}
\subfloat[\tiny mean+std saccade length]{
	\begin{tabular}{c c c }
		 \multicolumn{3}{c}{Prediction} \\ \hline
 		 unpleasant & neutral & pleasant  \\
		\cw{35.7} & \cb{34.0} & \cb{30.3} \\ 
		\cb{33.2} & \cw{31.6} & \cb{35.2} \\ 
		\cb{30.1} & \cb{34.4} & \cw{35.6} \\
		 \multicolumn{3}{c}{}\\
		 \multicolumn{3}{c}{mA=34.3\%}\\		
		 \multicolumn{3}{c}{($p=0.69$, 95\% CI= 30.85--37.80\%)}
	\end{tabular}
}
\subfloat[\tiny mean+std saccade slope]{
	\begin{tabular}{c c c }
		 \multicolumn{3}{c}{Prediction} \\ \hline
 		 unpleasant & neutral & pleasant  \\
		\cw{41.7} & \cb{32.7} & \cb{25.6} \\ 
		\cb{28.7} & \cw{34.0} & \cb{37.3} \\ 
		\cb{32.5} & \cb{33.0} & \cw{34.5} \\
		\multicolumn{3}{c}{}\\
		 \multicolumn{3}{c}{mA=37.16\%}\\		
		 \multicolumn{3}{c}{($p=0.31$, 95\% CI= 33.29--39.03\%)}
	\end{tabular}
}


\subfloat[\tiny histogram of fixation duration]{
	\begin{tabular}{c | l c c c }
		\multicolumn{1}{c}{} & & \multicolumn{3}{c}{Prediction} \\ \cline{3-5}
		\multicolumn{1}{c}{} & & unpleasant & neutral & pleasant \\
		\multirow{3}{*}{\rotatebox[origin=c]{90}{Actual}}		
		& unpleasant	& \cw{26.5} & \cb{38.5} & \cb{35.0} \\ 
		& neutral	& \cb{37.0} & \cw{27.6} & \cb{35.4} \\ 
		& pleasant	& \cb{36.5} & \cb{33.7} & \cw{29.8} \\
		\multicolumn{5}{r}{}\\
		\multicolumn{2}{r}{} & \multicolumn{3}{c}{mA=28.0\%}\\
		\multicolumn{2}{r}{} & \multicolumn{3}{c}{($p=0.05$, 95\% CI= 24.93--31.06\%)}
	\end{tabular}					
}
\subfloat[\tiny histogram of saccade length]{
	\begin{tabular}{c c c }
		 \multicolumn{3}{c}{Prediction} \\ \hline
 		 unpleasant & neutral & pleasant  \\
		\cw{35.3} & \cb{27.4} & \cb{37.2} \\ 
		\cb{33.5} & \cw{34.8} & \cb{31.7} \\ 
		\cb{31.2} & \cb{38.0} & \cw{30.8} \\
		 \multicolumn{3}{c}{}\\
		 \multicolumn{3}{c}{mA=33.5\%}\\		
		 \multicolumn{3}{c}{($p=0.94$, 95\% CI= 29.13--37.86\%)}
	\end{tabular}
}
\subfloat[\tiny histogram of saccade slope]{
	\begin{tabular}{c c c }
		 \multicolumn{3}{c}{Prediction} \\ \hline
 		 unpleasant & neutral & pleasant  \\
		\cw{33.3} & \cb{33.3} & \cb{23.3} \\ 
		\cb{33.3} & \cw{33.3} & \cb{33.3} \\ 
		\cb{33.3} & \cb{33.3} & \cw{33.3} \\		
		\multicolumn{3}{c}{}\\
		 \multicolumn{3}{c}{mA=33.3\%}\\		
		 \multicolumn{3}{c}{($p=1.00$, 95\% CI=33.33--33.33)}
	\end{tabular}
}


\subfloat[\tiny fixation density map]{
	\begin{tabular}{c | l c c c }
		\multicolumn{1}{c}{} & & \multicolumn{3}{c}{Prediction} \\ \cline{3-5}
		\multicolumn{1}{c}{} & & unpleasant & neutral & pleasant \\
		\multirow{3}{*}{\rotatebox[origin=c]{90}{Actual}}		
		& unpleasant	& \cw{33.0} & \cb{33.5} & \cb{33.5} \\ 
		& neutral	& \cb{35.2} & \cw{29.6} & \cb{35.2} \\ 
		& pleasant	& \cb{31.8} & \cb{36.9} & \cw{31.3} \\
		\multicolumn{5}{r}{}\\
		\multicolumn{2}{r}{} & \multicolumn{3}{c}{mA=31.3\%}\\
		\multicolumn{2}{r}{} & \multicolumn{3}{c}{($p=0.44$, 95\% CI=29.74--34.92)}
	\end{tabular}
}
\subfloat[\tiny IOVC]{
	\begin{tabular}{c c c }
		 \multicolumn{3}{c}{Prediction} \\ \hline
 		 unpleasant & neutral & pleasant  \\
		\cw{31.4} & \cb{34.5} & \cb{34.0} \\ 
		\cb{36.8} & \cw{30.9} & \cb{32.3} \\ 
		\cb{31.2} & \cb{34.9} & \cw{33.9} \\
		 \multicolumn{3}{c}{}\\
		 \multicolumn{3}{c}{mA=32.1\%}\\		
		 \multicolumn{3}{c}{($p=0.62$, 95\% CI= 28.49--35.50\%)}
	\end{tabular}
}
\subfloat[\tiny FKL]{
	\begin{tabular}{c c c }
		 \multicolumn{3}{c}{Prediction} \\ \hline
 		 unpleasant & neutral & pleasant  \\
		\cw{43.3} & \cb{31.4} & \cb{25.3} \\ 
		\cb{28.1} & \cw{33.2} & \cb{38.7} \\ 
		\cb{29.0} & \cb{35.3} & \cw{35.7} \\
		 \multicolumn{3}{c}{}\\
		 \multicolumn{3}{c}{mA=37.4\%}\\		
		 \multicolumn{3}{c}{($p=0.13$, 95\% CI= 29.78--44.87\%)}
	\end{tabular}
}
\end{tiny}
\caption{The performance of different eye movement features on the 95 images, where both genders highly agree on emotional gist of images and there is only one visual class category as described in~\cite{RezazadeganTavakoli2014}. Chance level is 33\%.}
\label{fig:confusion00}
\vskip -6pt
\end{table*}

\begin{figure*}
\centering
\begin{tiny}
\captionsetup[subfigure]{labelformat=empty}
\subfloat[\tiny mean+std fixation duration]{
	\begin{tabular}{c | l c c c }
		\multicolumn{1}{c}{} & & \multicolumn{3}{c}{Prediction} \\ \cline{3-5}
		\multicolumn{1}{c}{} & & unpleasant & neutral & pleasant \\
		\multirow{3}{*}{\rotatebox[origin=c]{90}{Actual}}		
		& unpleasant	& \cw{34.1} & \cb{33.4} & \cb{32.5} \\  
		& neutral	& \cb{33.5} & \cw{33.1} & \cb{33.5} \\ 
		& pleasant	& \cb{32.2} & \cb{33.5} & \cw{34.3} \\
		\multicolumn{5}{r}{}\\
		\multicolumn{2}{r}{} & \multicolumn{3}{c}{mA=33.8\%}\\
		\multicolumn{2}{r}{} & \multicolumn{3}{c}{($p=0.70$, 95\% CI=32.17--35.60)}
	\end{tabular}
}
\subfloat[\tiny mean+std saccade length]{
	\begin{tabular}{c c c }
		 \multicolumn{3}{c}{Prediction} \\ \hline
 		 unpleasant & neutral & pleasant  \\
		\cw{33.7} & \cb{33.5} & \cb{32.8} \\ 
		\cb{33.4} & \cw{32.6} & \cb{34.0} \\ 
		\cb{33.0} & \cb{33.7} & \cw{33.3} \\
		 \multicolumn{3}{c}{}\\
		 \multicolumn{3}{c}{mA=33.2\%}\\		
		 \multicolumn{3}{c}{($p=0.94$, 95\% CI= 32.23--34.21\%)}
	\end{tabular}
}
\subfloat[\tiny mean+std saccade slope]{
	\begin{tabular}{c c c }
		 \multicolumn{3}{c}{Prediction} \\ \hline
 		 unpleasant & neutral & pleasant  \\
		\cw{33.2} & \cb{33.2} & \cb{33.6} \\ 
		\cb{33.8} & \cw{33.0} & \cb{33.3} \\ 
		\cb{33.0} & \cb{33.8} & \cw{33.2} \\	
		\multicolumn{3}{c}{}\\
		 \multicolumn{3}{c}{mA=33.1\%}\\		
		 \multicolumn{3}{c}{($p=0.89$, 95\% CI= 31.65--34.56\%)}
	\end{tabular}
}


\subfloat[\tiny histogram of fixation duration]{
	\begin{tabular}{c | l c c c }
		\multicolumn{1}{c}{} & & \multicolumn{3}{c}{Prediction} \\ \cline{3-5}
		\multicolumn{1}{c}{} & & unpleasant & neutral & pleasant \\
		\multirow{3}{*}{\rotatebox[origin=c]{90}{Actual}}		
		& unpleasant	& \cw{31.8} & \cb{35.1} & \cb{33.1} \\ 
		& neutral	& \cb{32.9} & \cw{34.1} & \cb{32.9} \\ 
		& pleasant	& \cb{35.0} & \cb{31.0} & \cw{33.9} \\
		\multicolumn{5}{r}{}\\
		\multicolumn{2}{r}{} & \multicolumn{3}{c}{mA=33.4\%}\\
		\multicolumn{2}{r}{} & \multicolumn{3}{c}{($p=0.97$, 95\% CI= 30.48--36.29\%)}	
	\end{tabular}
}
\subfloat[\tiny histogram of saccade length]{
	\begin{tabular}{c c c }
		 \multicolumn{3}{c}{Prediction} \\ \hline
 		 unpleasant & neutral & pleasant  \\
		\cw{36.5} & \cb{33.4} & \cb{30.1} \\ 
		\cb{33.3} & \cw{30.4} & \cb{36.3} \\ 
		\cb{30.1} & \cb{36.0} & \cw{33.9} \\
		 \multicolumn{3}{c}{}\\
		 \multicolumn{3}{c}{mA=33.7\%}\\		
		 \multicolumn{3}{c}{($p=0.82$, 95\% CI= 30.48--36.29\%)}
	\end{tabular}
}
\subfloat[\tiny histogram of saccade slope]{
	\begin{tabular}{c c c }
		 \multicolumn{3}{c}{Prediction} \\ \hline
 		 unpleasant & neutral & pleasant  \\
		\cw{33.7} & \cb{37.0} & \cb{29.3} \\ 
		\cb{33.2} & \cw{32.7} & \cb{34.2} \\ 
		\cb{33.3} & \cb{31.9} & \cw{34.8} \\
		\multicolumn{3}{c}{}\\
		 \multicolumn{3}{c}{mA=33.7\%}\\		
		 \multicolumn{3}{c}{($p=0.87$, 95\% CI=32.12--35.10)}
	\end{tabular}
}


\subfloat[\tiny fixation density map]{
	\begin{tabular}{c | l c c c }
		\multicolumn{1}{c}{} & & \multicolumn{3}{c}{Prediction} \\ \cline{3-5}
		\multicolumn{1}{c}{} & & unpleasant & neutral & pleasant \\
		\multirow{3}{*}{\rotatebox[origin=c]{90}{Actual}}		
		& unpleasant	& \cw{42.0} & \cb{28.5} & \cb{29.5} \\ 
		& neutral	& \cb{28.8} & \cw{36.4} & \cb{34.8} \\ 
		& pleasant	& \cb{30.7} & \cb{34.2} & \cw{35.0} \\
		\multicolumn{5}{r}{}\\
		\multicolumn{2}{r}{} & \multicolumn{3}{c}{mA=37.8\%}\\
		\multicolumn{2}{r}{} & \multicolumn{3}{c}{($p=0.01$, 95\% CI=35.65--39.45)}
	\end{tabular}
}
\subfloat[\tiny IOVC]{
	\begin{tabular}{c c c }
		 \multicolumn{3}{c}{Prediction} \\ \hline
 		 unpleasant & neutral & pleasant  \\
		\cw{33.0} & \cb{34.8} & \cb{32.2} \\ 
		\cb{31.6} & \cw{34.1} & \cb{34.3} \\ 
		\cb{35.7} & \cb{31.2} & \cw{33.1} \\
		 \multicolumn{3}{c}{}\\
		 \multicolumn{3}{c}{mA=33.5\%}\\		
		 \multicolumn{3}{c}{($p=0.93$, 95\% CI= 32.01--34.90\%)}
	\end{tabular}
}
\subfloat[\tiny FKL]{
	\begin{tabular}{c c c }
		 \multicolumn{3}{c}{Prediction} \\ \hline
 		 unpleasant & neutral & pleasant  \\
		\cw{35.5} & \cb{34.6} & \cb{29.9} \\ 
		\cb{32.3} & \cw{33.7} & \cb{34.1} \\ 
		\cb{33.4} & \cb{32.2} & \cw{34.4} \\
		 \multicolumn{3}{c}{}\\
		 \multicolumn{3}{c}{mA=34.5\%}\\		
		 \multicolumn{3}{c}{($p=0.70$, 95\% CI= 32.17--35.61\%)}
	\end{tabular}
}

\end{tiny}
\caption{The performance of different eye movement features on the 296 images, where both genders highly agree on emotional gists of images, for recognizing the pleasantness of the images. Chance level is 33\%.}
\label{fig:confusion0}
\vskip -6pt
\end{figure*}

We repeated the experiments using the total of 382 images, which includes the images where genders do not agree. Figure~\ref{fig:confusion1} summarizes the results. We observe a change in the performance of features, that is, the fixation density map performance can not be distinguished from chance, while the histogram of saccade length starts showing some promising results with accuracy of 37.3\% [$p < 0.001$, 95\% CI=36.27--38.56]. It is worth mentioning that despite the performance of fixation density map is not distinguishable from chance on average, it provides a fair score for all the three classes.

\begin{figure*}
\centering
\begin{tiny}
\captionsetup[subfigure]{labelformat=empty}
\subfloat[\tiny mean+std fixation duration]{
	\begin{tabular}{c | l c c c }
		\multicolumn{1}{c}{} & & \multicolumn{3}{c}{Prediction} \\ \cline{3-5}
		\multicolumn{1}{c}{} & & unpleasant & neutral & pleasant \\
		\multirow{3}{*}{\rotatebox[origin=c]{90}{Actual}}		
		& unpleasant	& \cw{34.7} & \cb{32.5} & \cb{32.9} \\ 
		& neutral	& \cb{32.8} & \cw{33.3} & \cb{34.0} \\ 
		& pleasant	& \cb{32.4} & \cb{34.3} & \cw{33.4} \\		
		\multicolumn{5}{r}{}\\
		\multicolumn{2}{r}{} & \multicolumn{3}{c}{mA=33.8\%}\\
		\multicolumn{2}{r}{} & \multicolumn{3}{c}{($p=0.70$, 95\% CI=32.68--34.98)}
	\end{tabular}
}
\subfloat[\tiny mean+std saccade length]{
	\begin{tabular}{c c c }
		 \multicolumn{3}{c}{Prediction} \\ \hline
 		 unpleasant & neutral & pleasant  \\
		\cw{32.3} & \cb{35.3} & \cb{32.4} \\ 
		\cb{34.0} & \cw{32.5} & \cb{33.5} \\ 
		\cb{33.5} & \cb{32.4} & \cw{34.1} \\
		 \multicolumn{3}{c}{}\\
		 \multicolumn{3}{c}{mA=33.0\%}\\		
		 \multicolumn{3}{c}{($p=0.76$, 95\% CI= 31.19--34.64\%)}
	\end{tabular}
}
\subfloat[\tiny mean+std saccade slope]{
	\begin{tabular}{c c c }
		 \multicolumn{3}{c}{Prediction} \\ \hline
 		 unpleasant & neutral & pleasant  \\
		\cw{34.3} & \cb{32.8} & \cb{32.9} \\ 
		\cb{34.1} & \cw{33.0} & \cb{32.9} \\ 
		\cb{31.6} & \cb{34.2} & \cw{34.2} \\
		\multicolumn{3}{c}{}\\
		 \multicolumn{3}{c}{mA=33.8\%}\\		
		 \multicolumn{3}{c}{($p=0.71$, 95\% CI= 32.70--34.96\%)}
	\end{tabular}
}


\subfloat[\tiny histogram of fixation duration]{
	\begin{tabular}{c | l c c c }
		\multicolumn{1}{c}{} & & \multicolumn{3}{c}{Prediction} \\ \cline{3-5}
		\multicolumn{1}{c}{} & & unpleasant & neutral & pleasant \\
		\multirow{3}{*}{\rotatebox[origin=c]{90}{Actual}}		
		& unpleasant	& \cw{32.4} & \cb{32.7} & \cb{34.9} \\ 
		& neutral	& \cb{32.8} & \cw{34.0} & \cb{33.2} \\ 
		& pleasant	& \cb{34.7} & \cb{33.4} & \cw{32.0} \\
		\multicolumn{5}{r}{}\\
		\multicolumn{2}{r}{} & \multicolumn{3}{c}{mA=32.8\%}\\
		\multicolumn{2}{r}{} & \multicolumn{3}{c}{($p=0.66$, 95\% CI= 30.31--35.18\%)}
	\end{tabular}
}
\subfloat[\tiny histogram of saccade length]{
	\begin{tabular}{c c c }
		 \multicolumn{3}{c}{Prediction} \\ \hline
 		 unpleasant & neutral & pleasant  \\
		\cw{28.0} & \cb{34.9} & \cb{37.1} \\ 
		\cb{35.5} & \cw{35.0} & \cb{29.5} \\ 
		\cb{35.8} & \cb{30.3} & \cw{34.0} \\
		 \multicolumn{3}{c}{}\\
		 \multicolumn{3}{c}{mA=32.3\%}\\		
		 \multicolumn{3}{c}{($p=0.58$, 95\% CI= 30.93--34.15\%)}
	\end{tabular}
}
\subfloat[\tiny histogram of saccade slope]{
	\begin{tabular}{c c c }
		 \multicolumn{3}{c}{Prediction} \\ \hline
 		 unpleasant & neutral & pleasant  \\
		\cw{37.3} & \cb{34.6} & \cb{28.1} \\ 
		\cb{25.3} & \cw{31.6} & \cb{43.2} \\ 
		\cb{26.5} & \cb{30.4} & \cw{43.2} \\
		\multicolumn{3}{c}{}\\
		 \multicolumn{3}{c}{mA=37.3\%}\\		
		 \multicolumn{3}{c}{($p<0.001$, 95\% CI=36.27--38.56)}
	\end{tabular}
}


\subfloat[\tiny fixation density map]{
	\begin{tabular}{c | l c c c }
		\multicolumn{1}{c}{} & & \multicolumn{3}{c}{Prediction} \\ \cline{3-5}
		\multicolumn{1}{c}{} & & unpleasant & neutral & pleasant \\
		\multirow{3}{*}{\rotatebox[origin=c]{90}{Actual}}		
		& unpleasant	& \cw{35.2} & \cb{34.6} & \cb{30.1} \\ 
		& neutral	& \cb{31.7} & \cw{33.7} & \cb{34.6} \\ 
		& pleasant	& \cb{32.9} & \cb{31.4} & \cw{35.7} \\
		\multicolumn{5}{r}{}\\
		\multicolumn{2}{r}{} & \multicolumn{3}{c}{mA=34.9\%}\\
		\multicolumn{2}{r}{} & \multicolumn{3}{c}{($p=0.24$, 95\% CI=32.69--37.05)}
	\end{tabular}
}
\subfloat[\tiny IOVC]{
	\begin{tabular}{c c c }
		 \multicolumn{3}{c}{Prediction} \\ \hline
 		 unpleasant & neutral & pleasant  \\
\cw{33.9} & \cb{33.6} & \cb{32.5} \\ 
\cb{33.5} & \cw{33.5} & \cb{33.1} \\ 
\cb{32.6} & \cb{32.9} & \cw{34.4} \\
		 \multicolumn{3}{c}{}\\
		 \multicolumn{3}{c}{mA=33.9\%}\\		
		 \multicolumn{3}{c}{($p=0.34$, 95\% CI= 32.47--35.41\%)}
	\end{tabular}
}
\subfloat[\tiny FKL]{
	\begin{tabular}{c c c }
		 \multicolumn{3}{c}{Prediction} \\ \hline
 		 unpleasant & neutral & pleasant  \\
		 \cw{33.9} & \cb{29.7} & \cb{36.3} \\ 
		 \cb{32.1} & \cw{35.0} & \cb{32.9} \\ 
		 \cb{34.2} & \cb{33.8} & \cw{32.0} \\
		 \multicolumn{3}{c}{}\\
		 \multicolumn{3}{c}{mA=33.7\%}\\		
		 \multicolumn{3}{c}{($p=0.86$, 95\% CI= 30.67--36.48\%)}
	\end{tabular}
}

\end{tiny}
\caption{The performance of different eye movement features on the 382 images for recognizing the pleasantness of the images. In this set, there is no gender agreement on the emotional gist of some of the images and a diverse visual class category exist. Chance level is 33\%.}
\label{fig:confusion1}
\vskip -6pt
\end{figure*}

\emph{The combination of features} is also studied for all the three scenarios, where we perform a late fusion. We assessed the late fusion of all the eye movement features along with the fusion of fixation density map and the histogram of saccade slope, which seemed promising for the whole data set in individual feature experiments. The results are summarized in Figure~\ref{fig:confusion2}, depicting that none of the assessed combination settings does perform significantly above chance.

\begin{figure*}
\centering
\begin{tiny}
\captionsetup[subfigure]{labelformat=empty}
\subfloat[\tiny all features, 382 images]{
	\begin{tabular}{c | l c c c }
		\multicolumn{1}{c}{} & & \multicolumn{3}{c}{Prediction} \\ \cline{3-5}
		\multicolumn{1}{c}{} & & unpleasant & neutral & pleasant \\
		\multirow{3}{*}{\rotatebox[origin=c]{90}{Actual}}		
		& unpleasant	& \cw{37.1} & \cb{35.3} & \cb{27.5} \\ 
		& neutral	& \cb{30.2} & \cw{32.2} & \cb{37.6} \\ 
		& pleasant	& \cb{31.5} & \cb{31.8} & \cw{36.6} \\				
		\multicolumn{5}{r}{}\\
		\multicolumn{2}{r}{} & \multicolumn{3}{c}{mA=35.3\%}\\
		\multicolumn{2}{r}{} & \multicolumn{3}{c}{($p=0.44$, 95\% CI=32.53--36.12)}
	\end{tabular}
}
\subfloat[\tiny all features, 296 images]{
	\begin{tabular}{c c c }
		 \multicolumn{3}{c}{Prediction} \\ \hline
 		 unpleasant & neutral & pleasant  \\
		\cw{34.8} & \cb{33.0} & \cb{32.2} \\ 
		\cb{35.0} & \cw{32.4} & \cb{32.6} \\ 
		\cb{30.3} & \cb{34.5} & \cw{35.2} \\				
		 \multicolumn{3}{c}{}\\
		 \multicolumn{3}{c}{mA=34.2\%}\\		
		 \multicolumn{3}{c}{($p=0.59$, 95\% CI= 32.67-35.65\%)}
	\end{tabular}
}
\subfloat[\tiny all features, 95 images]{
	\begin{tabular}{c c c }
		 \multicolumn{3}{c}{Prediction} \\ \hline
 		 unpleasant & neutral & pleasant  \\
		\cw{31.4} & \cb{33.0} & \cb{35.6} \\ 
		\cb{34.7} & \cw{32.1} & \cb{33.2} \\ 
		\cb{33.8} & \cb{34.7} & \cw{31.5} \\
		\multicolumn{3}{c}{}\\
		 \multicolumn{3}{c}{mA=31.7\%}\\		
		 \multicolumn{3}{c}{($p=0.54$, 95\% CI= 27.85--35.47\%)}
	\end{tabular}
}


\subfloat[\tiny FDM + HSS, 382 images]{
	\begin{tabular}{c | l c c c }
		\multicolumn{1}{c}{} & & \multicolumn{3}{c}{Prediction} \\ \cline{3-5}
		\multicolumn{1}{c}{} & & unpleasant & neutral & pleasant \\
		\multirow{3}{*}{\rotatebox[origin=c]{90}{Actual}}		
		& unpleasant	& \cw{35.5} & \cb{35.8} & \cb{28.7} \\
		& neutral	& \cb{32.2} & \cw{30.5} & \cb{37.3} \\ 
		& pleasant	& \cb{29.9} & \cb{31.2} & \cw{38.9} \\
		\multicolumn{5}{r}{}\\
		\multicolumn{2}{r}{} & \multicolumn{3}{c}{mA=34.9\%}\\
		\multicolumn{2}{r}{} & \multicolumn{3}{c}{($p=0.16$, 95\% CI=33.05--36.86)}				
	\end{tabular}
}
\subfloat[\tiny FDM + HSS, 296 images]{
	\begin{tabular}{c c c }
		 \multicolumn{3}{c}{Prediction} \\ \hline
 		 unpleasant & neutral & pleasant  \\
		\cw{41.8} & \cb{27.4} & \cb{30.8} \\ 
		\cb{33.2} & \cw{33.5} & \cb{33.2} \\ 
		\cb{28.2} & \cb{36.8} & \cw{35.0} \\		
		 \multicolumn{3}{c}{}\\
		 \multicolumn{3}{c}{mA=36.8\%}\\		
		 \multicolumn{3}{c}{($p=0.09$, 95\% CI= 33.68--38.65\%)}
	\end{tabular}
}
\subfloat[\tiny FDM + HSS, 95 images]{
	\begin{tabular}{c c c }
		 \multicolumn{3}{c}{Prediction} \\ \hline
 		 unpleasant & neutral & pleasant  \\
		\cw{40.2} & \cb{30.4} & \cb{29.4} \\ 
		\cb{29.1} & \cw{32.9} & \cb{38.0} \\ 
		\cb{30.6} & \cb{37.2} & \cw{32.2} \\
		\multicolumn{3}{c}{}\\
		 \multicolumn{3}{c}{mA=35.1\%}\\		
		 \multicolumn{3}{c}{($p=0.48$, 95\% CI= 29.45--40.88\%)}
	\end{tabular}
}

\end{tiny}
\caption{The performance of the fusion of different eye movement-based features. The late fusion of all features and the combination of fixation density map (FDM) and the histogram of saccde slope (HSS) are presented. Chance level is 33\%.}
\label{fig:confusion2}
\vskip -6pt
\end{figure*}

In this experiment, while the results of the two settings achieving significantly better than chance performance consist with the findings that support the role of fixation patterns and saccade slope (angular behaviour) as reported in~\cite{Simola2015,Niu2012a}, we can not achieve a considerable classification score similar to those previously reported by~\cite{Soleymani2012} and our former research~\cite{RezazadeganTavakoli2014,R.-Tavakoli2015}. Apart from nuance feature and implementation differences, e.g.~\cite{Soleymani2012,R.-Tavakoli2015} employ some feature selection techniques, we believe our less restrictive experiment setup also plays a role and hence the current data is more difficult than the data used in earlier research. We will discuss this issue later.

\subsection{A Finer Look into the Fixation Locations}

We encoded fixation locations in terms of fixation density maps.
Among the features, by investigating the confusion matrices, the fixation density map seems performing fairly better than other features in two of the settings. It, particularly, performs significantly better than chance on the subset of 296 images. To summarize, the fixation density map provides the best performance where there exists diverse visual class categories and agreement on emotional gist of an image between genders. 
This gives rise to the following question, ``\emph{Why fixation density map does perform better on this specific setting?}''

The fixation density map can act as a holistic scene representation that provides a reasonable compressed  scene descriptor. This is basically the concept behind gist-based~\cite{Oliva2001} operators that exploit saliency for  scene recognition like~\cite{Siagian2007}. In other words, it seems the fixation density maps are helping to learn the visual class categories which can potentially be correlated with
the emotional gist of the scenes as some visual class categories are biased towards specific emotions, e.g. a photo of food is often identified as neutral or pleasant rather than unpleasant. This explains the reason behind the poor performance of the fixation density maps in the subset of 95 images, where there is only one visual class category available according to human-provided labels~\cite{RezazadeganTavakoli2014}. 

In the case of the 382 images, similarly, the performance is not significantly different from chance. We speculate this is associated with the effect of gender-bias, that is, some of the visual class categories are found emotionally different between the genders~\cite{Bradley2001,Gomez2013}, in 86 of these images, introduced by incorporating all the images. Such a bias will cause a visual class category to contribute to several emotional class categories and makes the pleasantness recognition from fixation density maps difficult.

We, thus, assess the performance of fixation density maps in terms of entropy to suppress the location information, i.e. we neutralize the role of visual class categories that may have been contributing to the recognition in the 296 images. Figure~\ref{fig:fdentropy} reports the results of the entropy of fixation density map, which is not significantly better than chance. In other words, discarding the spatial information, which holds a level of the visual category information of scenes, the detection of pleasantness becomes impossible even for the 296 images. Later, we will investigate the role of visual class categories with the help of visual features.

\begin{figure}
\centering
\begin{tiny}
\captionsetup[subfigure]{labelformat=empty}

\subfloat[\tiny entropy of fixation density map 95 images]{
	\begin{tabular}{c | l c c c }
		\multicolumn{1}{c}{} & & \multicolumn{3}{c}{Prediction} \\ \cline{3-5}
		\multicolumn{1}{c}{} & & unpleasant & neutral & pleasant \\
		\multirow{3}{*}{\rotatebox[origin=c]{90}{Actual}}		
		& unpleasant	& \cw{32.4} & \cb{31.0} & \cb{36.6} \\ 
		& neutral	& \cb{33.3} & \cw{33.3} & \cb{33.3} \\ 
		& pleasant	& \cb{33.9} & \cb{34.8} & \cw{31.3} \\
		\multicolumn{5}{r}{}\\
				\multicolumn{5}{r}{mA = 32.3}\\
		 \multicolumn{5}{r}{mA=32.3\% ($p=0.73$, 95\% CI=28.81--35.85\%) }\\						
	\end{tabular}
}

\subfloat[\tiny entropy of fixation density map 296 images]{
	\begin{tabular}{c | l c c c }
		\multicolumn{1}{c}{} & & \multicolumn{3}{c}{Prediction} \\ \cline{3-5}
		\multicolumn{1}{c}{} & & unpleasant & neutral & pleasant  \\
		\multirow{3}{*}{\rotatebox[origin=c]{90}{Actual}}		
		& unpleasant	& \cw{33.4} & \cb{33.1} & \cb{33.4} \\ 
		& neutral	& \cb{33.8} & \cw{32.8} & \cb{33.5} \\ 
		& pleasant	& \cb{32.5} & \cb{34.5} & \cw{33.0} \\		
		\multicolumn{5}{r}{}\\
		 \multicolumn{5}{r}{mA=33.1\% ($p=0.86$, 95\% CI=32.17--33.93\%) }\\						
	\end{tabular}
}

\subfloat[\tiny entropy of fixation density map 382 images]{
	\begin{tabular}{c | l c c c }
		\multicolumn{1}{c}{} & & \multicolumn{3}{c}{Prediction} \\ \cline{3-5}
		\multicolumn{1}{c}{} & & unpleasant & neutral & pleasant  \\
		\multirow{3}{*}{\rotatebox[origin=c]{90}{Actual}}		
		& unpleasant	& \cw{32.9} & \cb{32.6} & \cb{34.4} \\  
		& neutral	& \cb{33.1} & \cw{33.6} & \cb{33.3} \\ 
		& pleasant	& \cb{33.9} & \cb{33.9} & \cw{32.2} \\
		\multicolumn{5}{r}{}\\
		\multicolumn{5}{r}{mA=32.9\% ($p=0.73$, 95\% CI=30.09--34.81\%) }\\						
	\end{tabular}
}

\end{tiny}
\caption{Location independent fixation information: the performance of entropy of fixation density map. }
\label{fig:fdentropy}
\vskip -6pt
\end{figure}

\begin{figure}
\centering
\includegraphics[height=2.5in, width=3.2in]{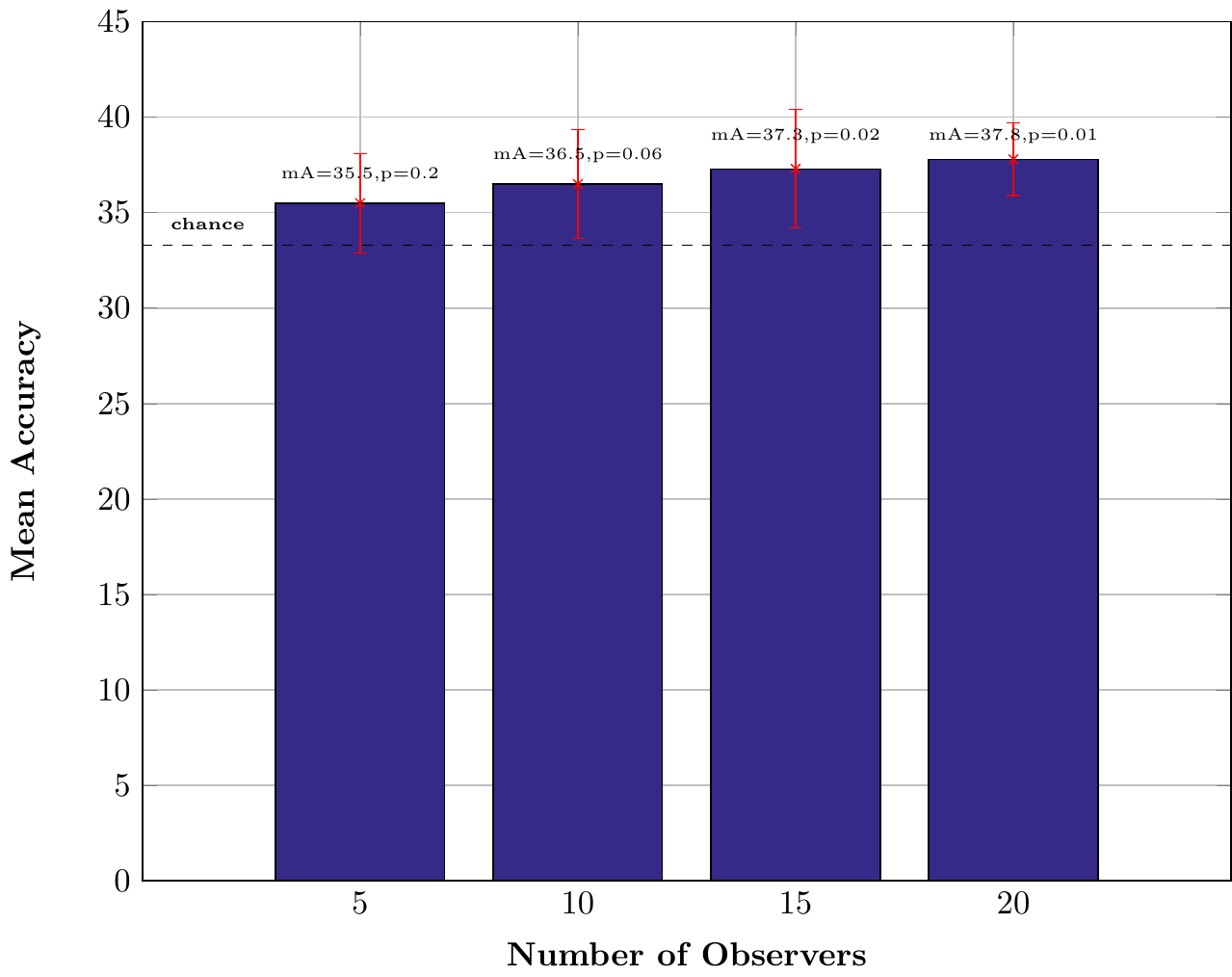}
\caption{The effect of number of observers on performance of fixation density map using the 296 images setup. }
\label{fig:fdmapObserver}
\vskip -6pt
\end{figure}

To further look into this phenomenon, we conduct an experiment by varying the number of observers and trying to predict the pleasantness for the 296 images. 
This is inspired by the observer consistency concept~\cite{Judd2012}, used for predicting the number of observers suitable for saliency prediction.
The hypothesis is that a minimum number of observers are needed to obtain a meaningful scene representation in order to achieve above chance performance in pleasantness recognition. 
We summarize the results in terms of the mean average accuracy and confidence intervals in Figure~\ref{fig:fdmapObserver}. It reveals that at least 15 observers are needed to achieve above chance classification.

\subsection{Visual Features}

We assessed the pleasantness detection from visual features in the three category (unpleasant, neutral, and pleasant) scenario. Similar to the experiments for eye movements, we use the three sets of images consisting of 95, 296, and 382 images. The visual features, which are of 4096 dimension, are L1-normalized and fed to the linear SVM. The results are summarized in Figure~\ref{fig:visfeat}.
In general, the average performance of visual features are in favor of deep features.

The PHOW features as reported in~\cite{RezazadeganTavakoli2014}  are not performing better than chance on any of the image sets. The performance for the deep features, however, is better than that of the PHOW features.  It is, however, worth noting that, considering the 95 images of~\cite{RezazadeganTavakoli2014}, which consists of only one visual class category of ``People and Daily Activity'', the performance is not better than chance for both features. 

By extending the image set to include more images, the deep features perform significantly above chance, that is 43.1\% [$p < 0.001$, 95\% CI=40.59--45.52] and 42.2\% [$p < 0.001$, 95\% CI=38.89--46.85], for 296 images and 382 images using VGG16, respectively. The results for VGG19 are also similar, 43.9\% [$p < 0.001$, 95\% CI=42.17--46.83] and 41.8\% [$p < 0.001$, 95\% CI=38.99--45.67] for 296 and 382 images.

Interestingly, the behaviour of deep features is similar to fixation density maps. We hypothesize that at least one reason behind such a performance behaviour potentially lies on the distribution of emotional categories and visual categories of the images, which is addressed in the following section.

\begin{figure*}
\centering
\begin{tiny}
\captionsetup[subfigure]{labelformat=empty}
\subfloat[\tiny PHOW]{
	\begin{tabular}{c | l c c c }
		\multicolumn{1}{c}{} & & \multicolumn{3}{c}{Prediction} \\ \cline{3-5}
		\multicolumn{1}{c}{} & & unpleasant & neutral & pleasant \\
		\multirow{3}{*}{\rotatebox[origin=c]{90}{Actual}}
		& unpleasant	& \cw{30.9} & \cb{30.9} & \cb{38.2} \\ 
		& neutral		& \cb{31.6} & \cw{34.7} & \cb{33.7} \\ 
		& pleasant		& \cb{38.0} & \cb{34.8} & \cw{27.2}\\
		\multicolumn{5}{c}{}\\
		\multicolumn{2}{r}{} &\multicolumn{3}{c}{mA=30.9\%}\\						
		\multicolumn{2}{r}{} &\multicolumn{3}{c}{($p=0.37$, 95\% CI= 28.03--34.98\%)}\\						
	\end{tabular}
		
}
\subfloat[\tiny VGG16 ]{
	\begin{tabular}{c c c }
		\multicolumn{3}{c}{Prediction} \\ \hline
      	   unpleasant & neutral & pleasant \\
		\cw{40.7} & \cb{28.9} & \cb{30.4} \\ 
		\cb{29.4} & \cw{36.5} & \cb{34.0} \\ 
		\cb{30.1} & \cb{34.4} & \cw{35.4} \\
		\multicolumn{3}{c}{}\\
  		\multicolumn{3}{c}{mA=37.6\% }\\
		 \multicolumn{3}{c}{($p=0.13$, 95\% CI=30.92--44.08\%) }\\						 
	\end{tabular}
}
\subfloat[\tiny VGG19]{
	\begin{tabular}{c c c c }
		\multicolumn{3}{c}{Prediction} \\ \cline{1-3}
      	   unpleasant & neutral & pleasant \\
		\cw{39.5} & \cb{30.8} & \cb{29.7} & \\ 
		\cb{32.8} & \cw{34.3} & \cb{32.8} & 95 images\\ 
		\cb{27.9} & \cb{34.8} & \cw{37.3} & \\
		\multicolumn{3}{c}{}\\
		\multicolumn{3}{c}{mA=37.0\% }\\
		 \multicolumn{3}{c}{($p=0.18$, 95\% CI=30.62--43.37\%) }\\						
	\end{tabular}
}

\captionsetup[subfigure]{labelformat=empty}
\subfloat[\tiny PHOW]{
	\begin{tabular}{c | l c c c }
		\multicolumn{1}{c}{} & & \multicolumn{3}{c}{Prediction} \\ \cline{3-5}
		\multicolumn{1}{c}{} & & unpleasant & neutral & pleasant \\
		\multirow{3}{*}{\rotatebox[origin=c]{90}{Actual}}
		& unpleasant	& \cw{36.1} & \cb{34.6} & \cb{29.3} \\ 
		& neutral		& \cb{31.7} & \cw{33.7} & \cb{34.6} \\ 
		& pleasant		& \cb{32.5} & \cb{31.9} & \cw{35.7} \\
		\multicolumn{5}{r}{}\\
		\multicolumn{2}{r}{} &\multicolumn{3}{c}{mA=35.1\%}\\
		\multicolumn{2}{r}{} &\multicolumn{3}{c}{ ($p=0.26$, 95\% CI=32.74--37.47\%) }\\						
	\end{tabular}
}
\subfloat[\tiny VGG16 ]{
	\begin{tabular}{c c c }
		\multicolumn{3}{c}{Prediction} \\ \hline
      	   unpleasant & neutral & pleasant \\
		\cw{49.9} & \cb{27.8} & \cb{22.3} \\ 
		\cb{21.9} & \cw{36.8} & \cb{41.3} \\ 
		\cb{22.7} & \cb{37.2} & \cw{40.1} \\
		\multicolumn{3}{c}{}\\
		\multicolumn{3}{c}{mA=43.1\%}\\
		\multicolumn{3}{c}{ ($p<0.001$, 95\% CI=40.59--45.52\%) }\\						
	\end{tabular}
}
\subfloat[\tiny VGG19]{
	\begin{tabular}{c c c c}
		\multicolumn{3}{c}{Prediction} \\  \cline{1-3}
      	   unpleasant  & neutral & pleasant\\
		\cw{47.1} & \cb{29.1} & \cb{23.8} \\ 
		\cb{27.6} & \cw{43.1} & \cb{29.3} & 296 images\\  
		\cb{25.0} & \cb{29.5} & \cw{45.5} \\
		\multicolumn{3}{c}{}\\
		\multicolumn{3}{c}{mA=43.9\%}\\
		\multicolumn{3}{c}{ ($p<0.001$, 95\% CI=42.17--46.83\%) }\\						
	\end{tabular}
}

\captionsetup[subfigure]{labelformat=empty}
\subfloat[\tiny PHOW]{
	\begin{tabular}{c | l c c c }
		\multicolumn{1}{c}{} & & \multicolumn{3}{c}{Prediction} \\ \cline{3-5}
		\multicolumn{1}{c}{} & & unpleasant  & neutral & pleasant \\
		\multirow{3}{*}{\rotatebox[origin=c]{90}{Actual}}
		& unpleasant	& \cw{34.0} & \cb{36.1} & \cb{29.9} \\ 
		& neutral	& \cb{32.4} & \cw{32.8} & \cb{34.8} \\ 
		& pleasant	& \cb{33.6} & \cb{31.4} & \cw{35.1} \\		
		\multicolumn{5}{c}{}\\
		 \multicolumn{2}{c}{}& \multicolumn{3}{c}{mA=34.0\% }\\						
 		 \multicolumn{2}{c}{}& \multicolumn{3}{c}{		 ($p=0.63$, 95\% CI=31.89--36.10\%) }\\						
	\end{tabular}
}
\subfloat[\tiny VGG16 ]{
	\begin{tabular}{c c c }
		\multicolumn{3}{c}{Prediction} \\ \hline
      	   unpleasant & neutral & pleasant  \\
		\cw{48.9} & \cb{28.0} & \cb{23.0} \\ 
		\cb{23.6} & \cw{37.1} & \cb{39.3} \\ 
		\cb{22.8} & \cb{36.5} & \cw{40.7} \\
		\multicolumn{3}{c}{}\\
		\multicolumn{3}{c}{mA=42.2\%}\\
		\multicolumn{3}{c}{ ($p<0.001$, 95\% CI=38.89--46.85\%) }\\						
	\end{tabular}
}
\subfloat[\tiny VGG19]{
	\begin{tabular}{c c c c }
		\multicolumn{3}{c}{Prediction} \\ \cline{1-3}
      	   unpleasant & neutral & pleasant  \\
		\cw{47.3} & \cb{28.6} & \cb{24.1} \\ 
		\cb{23.2} & \cw{37.8} & \cb{39.0}  & 382 images\\ 
		\cb{24.6} & \cb{35.2} & \cw{40.2} \\ 
		\multicolumn{3}{c}{}\\
		\multicolumn{3}{c}{mA=41.8\%}\\
		\multicolumn{3}{c}{ ($p<0.001$, 95\% CI=38.99--45.67\%) }\\						
	\end{tabular}
}
\end{tiny}
\caption{The performance of visual features for different number of images.}
\label{fig:visfeat}
\vskip -6pt
\end{figure*}

\subsection{Visual Categories}
\label{sec:vc}

In this experiment, we explicitly investigate the supporting evidence for the role of visual categories by using visual class category detectors to detect hidden visual categories of images. 
We apply an ImageNet~\cite{Deng2009} pre-trained model of object recognition~\cite{Simonyan2015} on the images to obtain the score for each of the 1000 classes of ImageNet making a feature vector of 1000 dimensions. 
Then, we train a classifier using the score of the visual class categories of each image as feature, akin to classemes~\cite{Torresani2010}. The results, summarized in Figure~\ref{fig:visclassdesc}, reveal that the visual class categories are strong pleasantness predictors when there exists enough visual class category diversity. Nonetheless, in the case of the 95 images, which consist of one hand-labelled class of ``People \& Daily Activity'', they fail performing significantly better than chance. This result consists with the performance of the fixation density maps and their role as visual scene descriptors.

\begin{figure}
\centering
\subfloat[95 images, 61 detected ImageNet visual categories]{
\includegraphics[height=1.8in, width=2.9in]{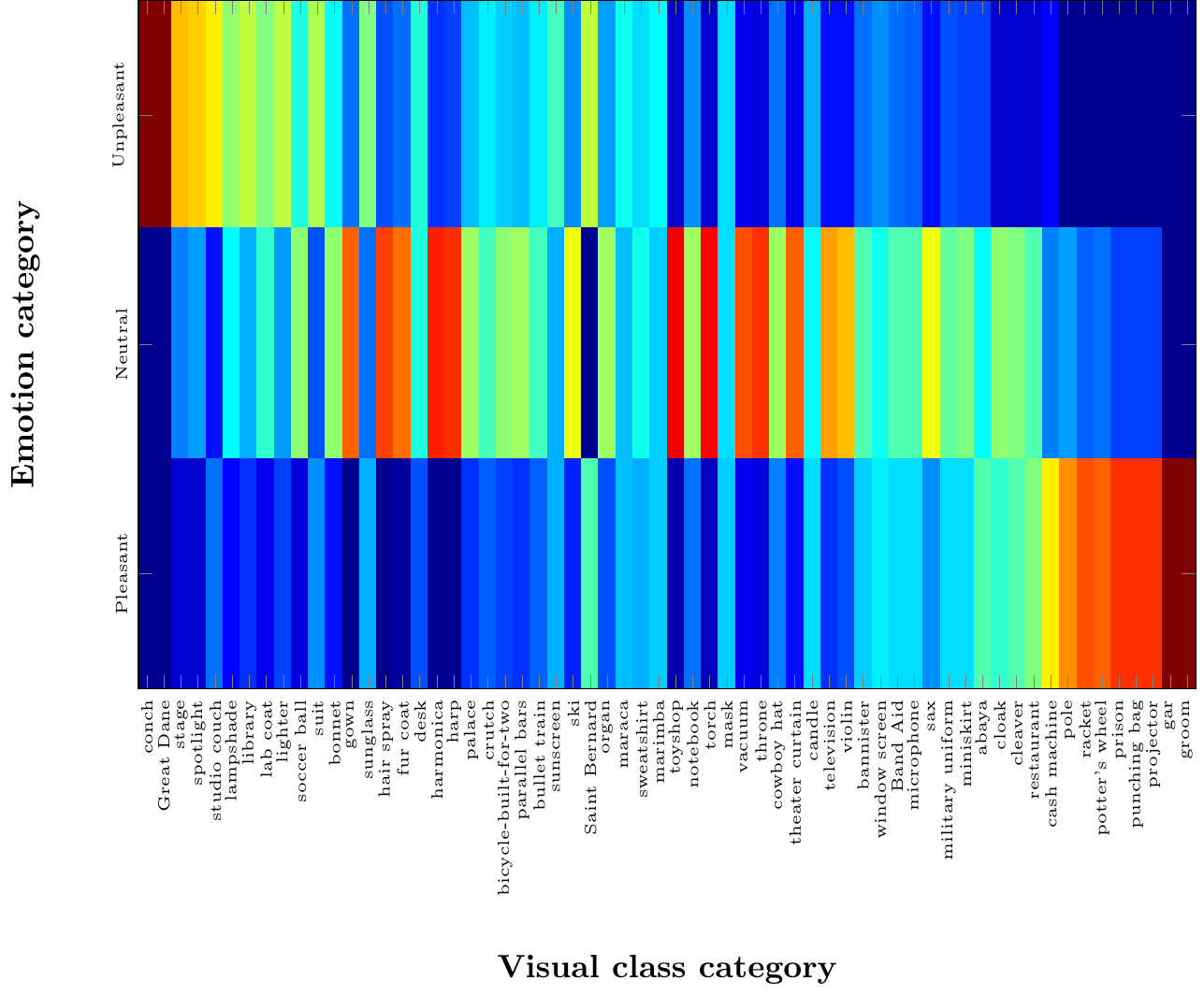}}

\subfloat[296 images, 202 detected ImageNet visual categories]{
\includegraphics[height=1.8in, width=2.9in]{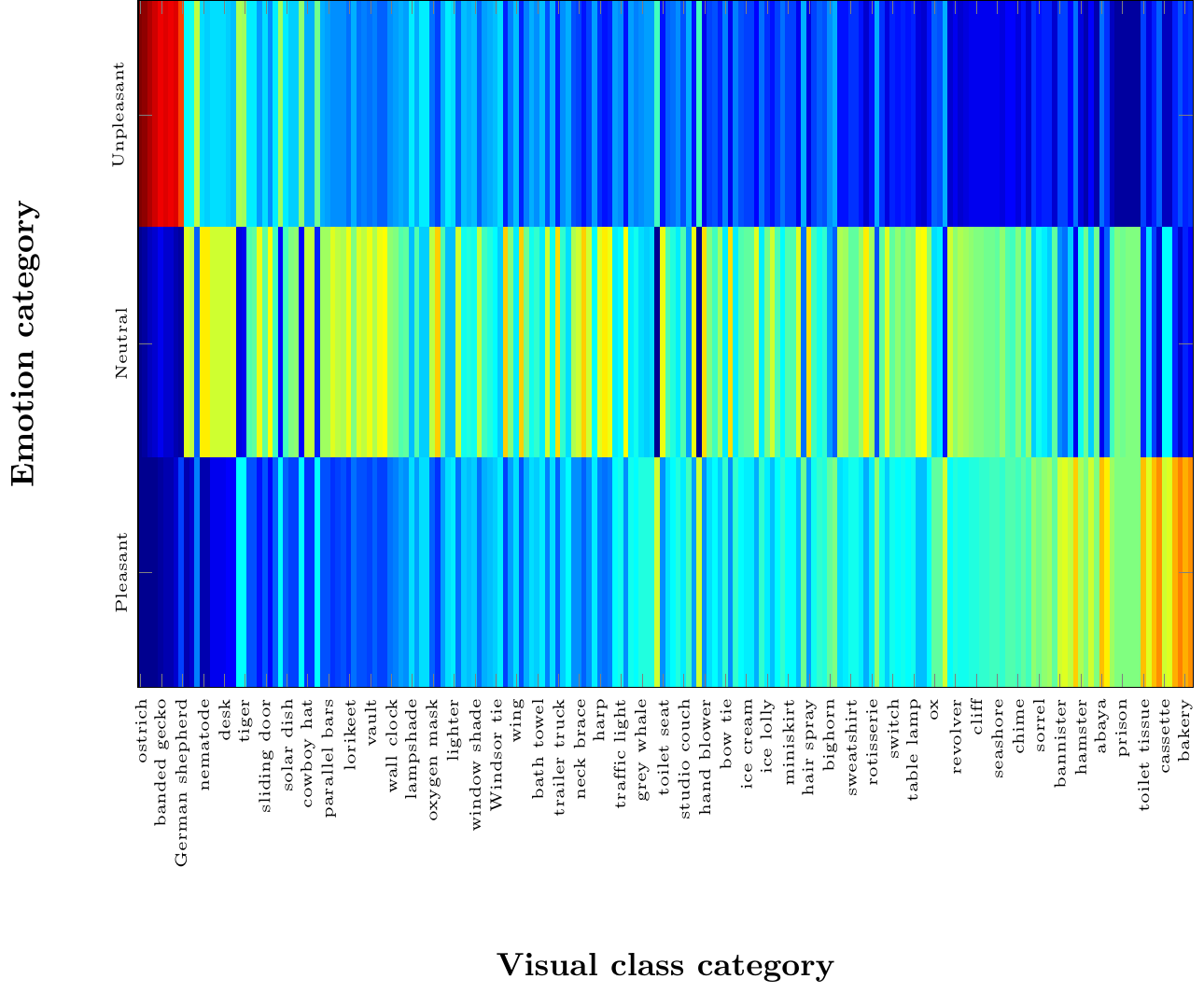}}

\subfloat[382 images, 240 detected ImageNet visual categories]{
\includegraphics[height=1.8in, width=2.9in]{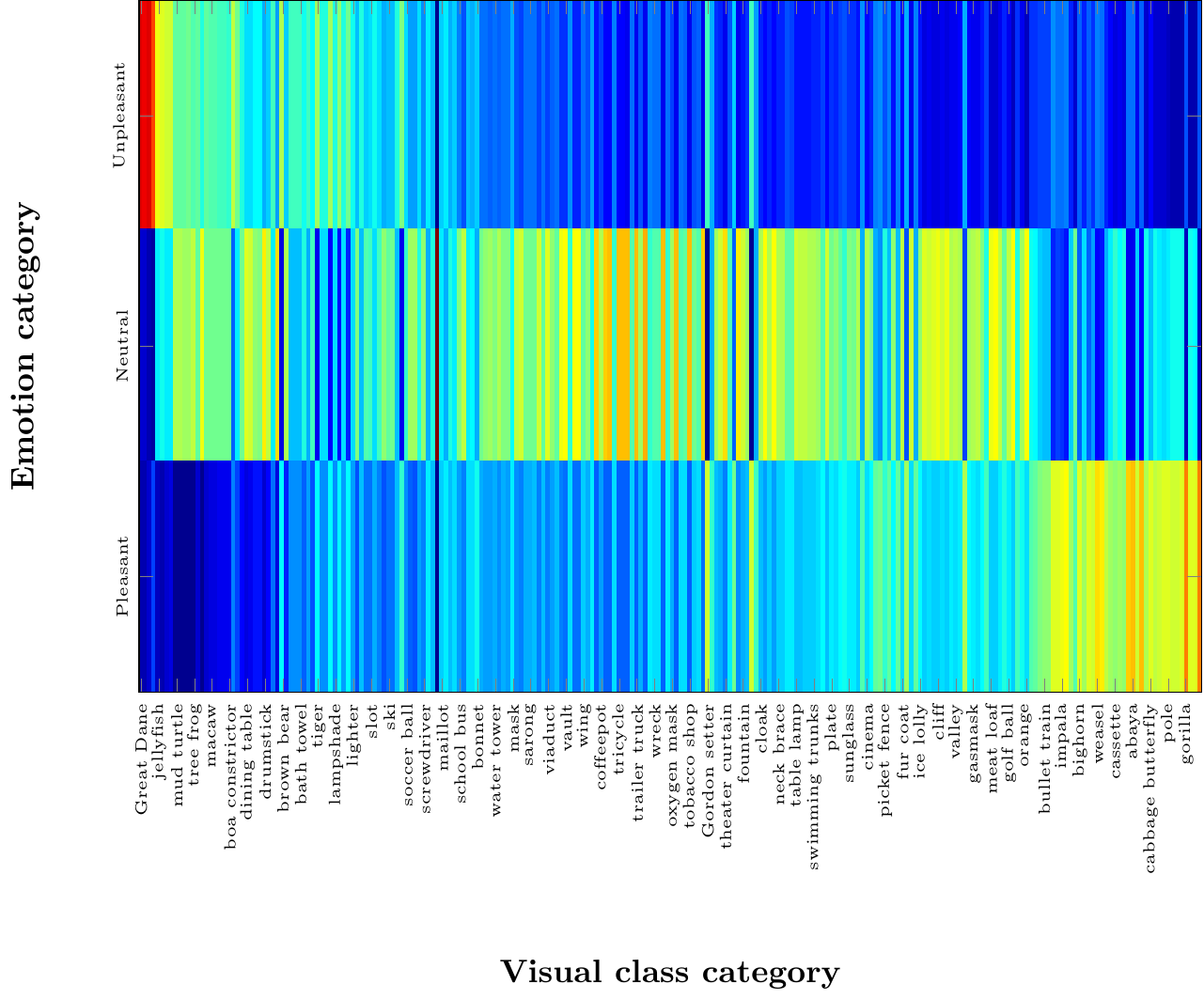}}

\caption{Visual class categories detected with the VGG16 network and their valance according to emotional ratings. It seems there is a strong bias in visual class categories, towards a specific valence value. Some class categories are more pleasant and some are more unpleasant, while some class categories are having multiple ratings. The red color indicates higher probability. From top to bottom, the image emotion changes from unpleasant to pleasant. Selected visual class names are shown for better visualization.}
\label{fig:classEmotion}
\vskip -6pt
\end{figure}

To further look into the role of the visual class categories, by assigning the visual class label of the maximum score to each image, we can detect 61 visual classes within 95 images, 202 visual class categories within the 296 images, and 240 visual classes within 382 images. We visualize the detected ImageNet visual category and their ground truth emotional class category in Figure~\ref{fig:classEmotion}. These visualizations clearly indicate that some visual classes are emotionally different from each other and detecting visual categories can facilitate emotion prediction as there seems to be a strong bias towards different emotions in the visual categories. For example, based on this analysis on 382 images, ``cock'' is absolutely neutral, while a ``tarantula'' is definitely unpleasant. There are also visual class categories of mixed valence such as ``maillot'', which is either neutral or pleasant. This latter example also signifies the role of agreement on pleasantness, as the visual class category of ``maillot'' does not exist in the 296 images where both visual features and fixation density maps perform best.

\begin{figure}
\centering
\begin{tiny}
\subfloat[95 images]{
	\begin{tabular}{c | l c c c }
		\multicolumn{1}{c}{} & & \multicolumn{3}{c}{Prediction} \\ \cline{3-5}
		\multicolumn{1}{c}{} & & unpleasant & neutral & pleasant  \\
		\multirow{3}{*}{\rotatebox[origin=c]{90}{Actual}}
		& unpleasant	& \cw{32.0} & \cb{30.1} & \cb{37.9} \\ 
		& neutral	& \cb{33.0} & \cw{33.9} & \cb{33.0} \\ 
		& pleasant	& \cb{34.5} & \cb{35.0} & \cw{30.5} \\		
		\multicolumn{5}{r}{}\\			
		\multicolumn{5}{c}{mA=32.1\%, ($p=0.68$, 95\% CI=26.54--37.79\%) }						
	\end{tabular}
}

\subfloat[296 images]{
	\begin{tabular}{c | l c c c }
		\multicolumn{1}{c}{} & & \multicolumn{3}{c}{Prediction} \\ \cline{3-5}
		\multicolumn{1}{c}{} & & unpleasant & neutral & pleasant  \\
		\multirow{3}{*}{\rotatebox[origin=c]{90}{Actual}}
		& unpleasant	& \cw{50.8} & \cb{28.2} & \cb{21.0} \\ 
		& neutral	& \cb{25.0} & \cw{38.8} & \cb{36.2} \\ 
		& pleasant	& \cb{26.0} & \cb{32.2} & \cw{41.7} \\		
		\multicolumn{5}{r}{}\\			
		\multicolumn{5}{c}{mA=43.5\%, ($p<0.001$, 95\% CI=40.09--46.91\%) }						
	\end{tabular}
}

\subfloat[382 images]{
	\begin{tabular}{c | l c c c }
		\multicolumn{1}{c}{} & & \multicolumn{3}{c}{Prediction} \\ \cline{3-5}
		\multicolumn{1}{c}{} & & unpleasant & neutral & pleasant  \\
		\multirow{3}{*}{\rotatebox[origin=c]{90}{Actual}}
		& unpleasant	& \cw{52.4} & \cb{28.2} & \cb{19.4} \\ 
		& neutral	& \cb{21.5} & \cw{38.0} & \cb{40.5} \\ 
		& pleasant	& \cb{27.1} & \cb{33.7} & \cw{39.2} \\		
		\multicolumn{5}{r}{}\\			
		\multicolumn{5}{r}{mA=43.2\%, ($p<0.001$, 95\% CI=40.44--45.56\%) }						
	\end{tabular}
}
\end{tiny}
\caption{The performance of detection from class categories using VGG16 detections:  confusion matrix.}
\label{fig:visclassdesc}
\vskip -6pt
\end{figure}

\subsection{Eye Movements and Deep Features}

To assess the potential gain obtained by combining  the eye movement and deep image features, we study the combination of eye movements and deep visual features in a late fusion scheme. The following combinations are studied: the visual features and fixation density maps, and the visual features and all eye movement features. 
For the 296 images, the obtained mean accuracy is $43.3\%$ [$p<0.001$, 95\% CI=41.3--46.68] and 39.8\% [$p<0.001$, 95\% CI=37.55--42.66], respectively. The confusion matrices are summarized in Figure~\ref{fig:visfeatEye}. Similar results hold for the 382 images, where the mean accuracy is $42.5\%$ [$p<0.001$, 95\% CI=40.2--46.11] and 39.3\% [$p<0.001$, 95\% CI=37.51--41.31] for visual features and fixation density maps, and visual features and all eye movement features, respectively. Since the eye movement signals of the current data are not that informative in our framework, it is not surprising that the combination of features also does not improve much over the deep features.

\begin{figure}
\centering
\begin{tiny}

\captionsetup[subfigure]{labelformat=empty}
\subfloat[\tiny fixation density map + VGG16 ]{
	\begin{tabular}{c | l c c c }
		\multicolumn{1}{c}{} & & \multicolumn{3}{c}{Prediction} \\ \cline{3-5}
		\multicolumn{1}{c}{} & & unpleasant & neutral & pleasant  \\
		\multirow{3}{*}{\rotatebox[origin=c]{90}{Actual}}
		& unpleasant	& 		\cw{51.3} & \cb{26.1} & \cb{22.6} \\ 
		& neutral 		& \cb{20.6} & \cw{38.8} & \cb{40.5} \\ 
		& pleasant		& \cb{23.0} & \cb{37.1} & \cw{39.8} \\
		 \multicolumn{5}{r}{}\\
		 \multicolumn{5}{r}{mA=43.3\% ($\kappa=0.37$, 95\% CI=41.3--46.68\%)}\\		
	\end{tabular}
}

\subfloat[\tiny all eye features + VGG16]{
	\begin{tabular}{c | l c c c }
		\multicolumn{1}{c}{} & & \multicolumn{3}{c}{Prediction} \\ \cline{3-5}
		\multicolumn{1}{c}{} & & unpleasant & neutral& pleasant  \\
		\multirow{3}{*}{\rotatebox[origin=c]{90}{Actual}}
		& unpleasant	&  \cw{44.2} & \cb{30.4} & \cb{25.4} \\  
		& neutral 		& \cb{27.2} & \cw{36.1} & \cb{36.7} \\  
		& pleasant		& \cb{26.7} & \cb{34.1} & \cw{39.2} \\
		 \multicolumn{5}{r}{}\\
		 \multicolumn{5}{r}{mA=39.8\% ($p<0.001$, 95\% CI=37.55--42.66\%)}\\	
	\end{tabular}
}
\end{tiny}
\caption{The combination of visual features and eye movements for the 296 images:  confusion matrices.}
\label{fig:visfeatEye}
\vskip -6pt
\end{figure}

\section{Discussion}

Affective multimedia tagging is a challenging task. It becomes substantially difficult when dealing with images rather than videos.
In the case of videos, the audio combined with the visual features make a strong cue for determining the emotional message of a scene. On the other hand, the affective tagging of a still image mostly relies only on visual features and the physiological signals of the observers. In this study, we focused on visual features and eye movements as cues for detecting the pleasantness of still images. The emphasis, however, is on the recording of the eye movements mimicking a setup of everyday user experience using an affordable eye tracking device.
Consequently, the current data is difficult because it potentially involves a higher level of noise compared with the previous datasets. 

The proposed approach and features from eye movements are able to decode image pleasantness only in one setup. The experiment arrangement plays a major role in this respect because we are using a less restrictive setup. Compared to a controlled setup, we are not using a dim room and do not use a chin rest, while the user performs calibration himself/herself after receiving training, consistent with the HCI concept of affordable eye tracking device development. We are using an affordable eye tracking device that has a minimum sampling rate of 60Hz. Not having any control over selecting a fixed sampling rate, makes the precision of fixation durations, and any property relating to time, vulnerable to noise. It is worth noting that we are using the API provided by the manufacturer to obtain eye movement features, e.g., fixations, fixation durations, and gaze points. This can be alerting in the sense that scaling up the utilization of the inference from eye movement using current affordable technologies might be challenging if not impossible. Furthermore, we also tried optimizing the feature parameters, e.g., the number of bins, etc. via cross-validation. Nonetheless, the results were not significantly different for various parameter values.

On the use of computer-vision-based techniques, we focused mostly on techniques that analyze the image content rather than the users' facial expressions. While the latter potentially provide a better user-dependent emotion estimation, that can be aggregated to infer a user-independent majority vote tag, we would like to rely on techniques that facilitate easier concealing of the observer identity  and provide a means of unobtrusive signal crowd-sourcing. Thus, facial expression analysis was avoided in this study. It is worth noting that although an eye tracker uses a camera, the hardware can be tweaked not to keep track of the faces, which alleviates the privacy concerns.

In this study, we found that visual categories are a strong predictor for user-independent pleasantness recognition of images.  This finding is indeed well aligned with the studies promoting that image class categories are emotionally different from each other~\cite{Bradley2007a}. It necessitates us to revisit the computer vision techniques, particularly those developed by adapting pre-trained deep CNNs of image classification tasks like~\cite{You2015,Wang2016}, and to study the emotional bias of visual class categories in the databases. Unfortunately, such a study goes beyond the current manuscript. We, thus, will address it separately in future work.

The successful application of eye movements as the sole inference medium depends on many factors including the sensor, the experiment setup and the employed machine learning techniques. Using only the data from an affordable eye tracker, we could fairly well decode the image pleasantness in a subset of images. Considering our less constrained setup, which resembles a scenario in which a naive user interacts with a computer freely, we are slightly alerted that the wide-spread utilization of some signals, such as eye movements, may not be as easy as the community may have expected.  Despite some improvements, e.g.~\cite{Zhang2015,Borji2014a,Kanan2014}, we still need robust and vigorous machine learning techniques and more reliable sensors. To conclude, it seems that inference from eye movements is difficult using current sensors and techniques. The success of a method can also be severly affected by the level of control in the interaction and experiment, which is alerting for HCI interfaces in daily use. Even in controlled experiments, it is worthnoting that the controversy may exist (e.g., check~\cite{Greene2012} versus~\cite{Haji-Abolhassani2014}). 

The data set is available at~\url{https://github.com/hrtavakoli/IMPEDE}.

\section{Conclusions}

In this study, we investigated the performance of eye movements for valence recognition in natural images in a less restrictive setup by introducing a new data set. An affordable eye tracker (more precisely a Tobii EyeX controller) was put into test and several approaches were studied. The results are promising in one specific case, albeit not compelling enough. This alerts us to be more careful while hoping the wide-spread use of eye tracking-based inferences in the HCI paradigm soon, at least for image pleasantness recognition.

We also studied visual features based on the deep CNNs. While the traditional features failed for pleasantness detection using natural images, deep CNNs were outperforming all the features and provided the top performance in our experiments.  Evaluating the visual features further, we revealed that the visual class categories are indeed strong valence predictors. Although such a phenomenon can help boosting the algorithms that are based on visual features, it also necessitates revisiting the results of the existing methods in a bias-free setting.

\section{Acknowledgement}
This work was supported by the Finnish Center of Excellence in Computational Inference Research (COIN).

\section*{References}

\bibliography{valenceBiblo}

\begin{thebibliography}{10}
\expandafter\ifx\csname url\endcsname\relax
  \def\url#1{\texttt{#1}}\fi
\expandafter\ifx\csname urlprefix\endcsname\relax\def\urlprefix{URL }\fi
\expandafter\ifx\csname href\endcsname\relax
  \def\href#1#2{#2} \def\path#1{#1}\fi

\bibitem{Barral2015}
O.~Barral, M.~J. Eugster, T.~Ruotsalo, M.~M. Spap{\'e}, I.~Kosunen, N.~Ravaja,
  S.~Kaski, G.~Jacucci, Exploring peripheral physiology as a predictor of
  perceived relevance in information retrieval, in: Proceedings of the 20th
  International Conference on Intelligent User Interfaces, IUI '15, ACM, New
  York, NY, USA, 2015, pp. 389--399.
\newblock \href {http://dx.doi.org/10.1145/2678025.2701389}
  {\path{doi:10.1145/2678025.2701389}}.

\bibitem{Yanulevskaya2008}
V.~Yanulevskaya, J.~van Gemert, K.~Roth, A.~Herbold, N.~Sebe, J.~Geusebroek,
  Emotional valence categorization using holistic image features, in: Image
  Processing, 2008. ICIP 2008. 15th IEEE International Conference on, 2008, pp.
  101 --104.
\newblock \href {http://dx.doi.org/10.1109/ICIP.2008.4711701}
  {\path{doi:10.1109/ICIP.2008.4711701}}.

\bibitem{Machajdik2010}
J.~Machajdik, A.~Hanbury, Affective image classification using features
  inspired by psychology and art theory, in: Proceedings of the international
  conference on Multimedia, MM '10, ACM, New York, NY, USA, 2010, pp. 83--92.

\bibitem{Cho2004}
S.-B. Cho, Emotional image and musical information retrieval with interactive
  genetic algorithm, Proceedings of the IEEE 92~(4) (2004) 702--711.
\newblock \href {http://dx.doi.org/10.1109/JPROC.2004.825900}
  {\path{doi:10.1109/JPROC.2004.825900}}.

\bibitem{Wang2008}
W.~Wang, Q.~He, A survey on emotional semantic image retrieval, in: 15th IEEE
  Int. Conf. on Image Processing, 2008, pp. 117--120.

\bibitem{RezazadeganTavakoli2014}
H.~Tavakoli, V.~Yanulevskaya, E.~Rahtu, J.~Heikkila, N.~Sebe, Emotional valence
  recognition, analysis of salience and eye movements, in: ICPR, 2014.

\bibitem{Itten1973}
J.~Itten, The art of color : the subjective experience and objective rationale
  of color, John Wiley, New York, 1973.

\bibitem{Colombo1999}
C.~Colombo, A.~Del~Bimbo, P.~Pala, Semantics in visual information retrieval,
  Multimedia, IEEE 6~(3) (1999) 38 --53.
\newblock \href {http://dx.doi.org/10.1109/93.790610}
  {\path{doi:10.1109/93.790610}}.

\bibitem{Ou2004}
L.-C. Ou, M.~R. Luo, A.~Woodcock, A.~Wright, A study of colour emotion and
  colour preference. {P}art {I}: Colour emotions for single colours, Color
  Research \& Application 29~(3) (2004) 232--240.
\newblock \href {http://dx.doi.org/10.1002/col.20010}
  {\path{doi:10.1002/col.20010}}.

\bibitem{Lang2008}
P.~Lang, M.~Bradley, B.~Cuthbert, International affective picture system
  ({IAPS}): Affective ratings of pictures and instruction manual, Tech. Rep.
  A-8, University of Florida, Gainesville, FL (2008).

\bibitem{Lang1995}
P.~Lang, The emotion probe: Studies of motivation and attention, American
  psychologist 50 (1995) 372--372.

\bibitem{Bradley2007}
M.~M. Bradley, S.~Hamby, A.~Löw, P.~J. Lang, Brain potentials in perception:
  Picture complexity and emotional arousal, Psychophysiology 44~(3) (2007)
  364--373.
\newblock \href {http://dx.doi.org/10.1111/j.1469-8986.2007.00520.x}
  {\path{doi:10.1111/j.1469-8986.2007.00520.x}}.

\bibitem{Bradley2001}
M.~M. Bradley, M.~Codispoti, D.~Sabatinelli, P.~Lang, Emotion and motivation
  {II}: sex differences in picture processing, Emotion 1~(3) (2001) 276--298.

\bibitem{Arnheim1974}
R.~Arnheim, Art and visual perception: A psychology of the creative eye, 1974.

\bibitem{Bar2006}
M.~Bar, M.~Neta, Humans prefer curved visual objects, Psychological Science
  17~(8) (2006) 645--648.

\bibitem{Nummenmaa2006}
L.~Nummenmaa, J.~Hy{\"o}n{\"a}, M.~G. Calvo, Eye movement assessment of
  selective attentional capture by emotional pictures, Emotion 6~(2) (2006)
  257--268.

\bibitem{Humphrey2012}
K.~Humphrey, G.~Underwood, T.~Lambert, Salience of the lambs: A test of the
  saliency map hypothesis with pictures of emotive objects, Journal of Vision
  12~(1).
\newblock \href
  {http://arxiv.org/abs/http://www.journalofvision.org/content/12/1/22.full.pdf+html}
  {\path{arXiv:http://www.journalofvision.org/content/12/1/22.full.pdf+html}},
  \href {http://dx.doi.org/10.1167/12.1.22} {\path{doi:10.1167/12.1.22}}.

\bibitem{Niu2012a}
Y.~Niu, R.~M. Todd, M.~Kyan, A.~K. Anderson, Visual and emotional salience
  influence eye movements, ACM Trans. Appl. Percept. 9~(3) (2012) 13:1--13:18.
\newblock \href {http://dx.doi.org/10.1145/2325722.2325726}
  {\path{doi:10.1145/2325722.2325726}}.

\bibitem{Wang2005}
W.-N. Wang, Y.-L. Yu, Image emotional semantic query based on color semantic
  description, in: Machine Learning and Cybernetics, 2005. Proceedings of 2005
  International Conference on, Vol.~7, 2005, pp. 4571--4576 Vol. 7.

\bibitem{Solli2008}
M.~Solli, R.~Lenz, Color emotions for image classification and retrieval, in:
  CGIV, 2008, p. 367–371.

\bibitem{Solli2009}
M.~Solli, R.~Lenz, Color based bags-of-emotions, in: Proceedings of the 13th
  International Conference on Computer Analysis of Images and Patterns, CAIP
  '09, Springer-Verlag, Berlin, Heidelberg, 2009, pp. 573--580.

\bibitem{Yanulevskaya2012}
V.~Yanulevskaya, J.~Uijlings, E.~Bruni, A.~Sartori, E.~Zamboni, F.~Bacci,
  D.~Melcher, N.~Sebe, In the eye of the beholder: employing statistical
  analysis and eye tracking for analyzing abstract paintings, in: Proceedings
  of the 20th ACM international conference on Multimedia, MM '12, ACM, New
  York, NY, USA, 2012, pp. 349--358.

\bibitem{Sartori2015}
A.~Sartori, D.~Culibrk, Y.~Yan, N.~Sebe, Who's afraid of {I}tten: Using the art
  theory of color combination to analyze emotions in abstract paintings, in:
  Proceedings of the 23rd ACM International Conference on Multimedia, 2015, pp.
  311--320.

\bibitem{Bovik1990}
A.~Bovik, M.~Clark, W.~Geisler, Multichannel texture analysis using localized
  spatial filters, Pattern Analysis and Machine Intelligence, IEEE Transactions
  on 12~(1) (1990) 55--73.
\newblock \href {http://dx.doi.org/10.1109/34.41384}
  {\path{doi:10.1109/34.41384}}.

\bibitem{Vailaya2001}
A.~Vailaya, M.~A. Figueiredo, A.~K. Jain, H.-J. Zhang, Image classification for
  content-based indexing, Trans. Img. Proc. 10~(1) (2001) 117--130.
\newblock \href {http://dx.doi.org/10.1109/83.892448}
  {\path{doi:10.1109/83.892448}}.

\bibitem{Lu2012}
X.~Lu, P.~Suryanarayan, R.~B. Adams, Jr., J.~Li, M.~G. Newman, J.~Z. Wang, On
  shape and the computability of emotions, in: Proceedings of the 20th ACM
  international conference on Multimedia, MM '12, ACM, New York, NY, USA, 2012,
  pp. 229--238.
\newblock \href {http://dx.doi.org/10.1145/2393347.2393384}
  {\path{doi:10.1145/2393347.2393384}}.

\bibitem{Lew2006}
M.~S. Lew, N.~Sebe, C.~Djeraba, R.~Jain, Content-based multimedia information
  retrieval: State of the art and challenges, ACM Trans. Multimedia Comput.
  Commun. Appl. 2~(1) (2006) 1--19.
\newblock \href {http://dx.doi.org/10.1145/1126004.1126005}
  {\path{doi:10.1145/1126004.1126005}}.

\bibitem{Krizhevsky2012}
A.~Krizhevsky, I.~Sutskever, G.~E. Hinton, Imagenet classification with deep
  convolutional neural networks, in: F.~Pereira, C.~Burges, L.~Bottou,
  K.~Weinberger (Eds.), Advances in Neural Information Processing Systems 25,
  Curran Associates, Inc., 2012, pp. 1097--1105.

\bibitem{Koskela2014}
M.~Koskela, J.~Laaksonen, Convolutional network features for scene recognition,
  in: Proceedings of the 22Nd ACM International Conference on Multimedia, MM
  '14, ACM, New York, NY, USA, 2014, pp. 1169--1172.
\newblock \href {http://dx.doi.org/10.1145/2647868.2655024}
  {\path{doi:10.1145/2647868.2655024}}.

\bibitem{Szegedy2013}
C.~Szegedy, A.~Toshev, D.~Erhan, Deep neural networks for object detection, in:
  C.~Burges, L.~Bottou, M.~Welling, Z.~Ghahramani, K.~Weinberger (Eds.),
  Advances in Neural Information Processing Systems 26, 2013, pp. 2553--2561.

\bibitem{Long2015}
J.~Long, E.~Shelhamer, T.~Darrell, Fully convolutional networks for semantic
  segmentation, CVPR (to appear)\href {http://arxiv.org/abs/1411.4038}
  {\path{arXiv:1411.4038}}.

\bibitem{You2015}
Q.~You, J.~Luo, H.~Jin, J.~Yang, Robust image sentiment analysis using
  progressively trained and domain transferred deep networks, in: Proceedings
  of the Twenty-Ninth AAAI Conference on Artificial Intelligence, AAAI'15, AAAI
  Press, 2015, pp. 381--388.

\bibitem{Wang2016}
J.~Wang, J.~Fu, Y.~Xu, T.~Mei, Beyond object recognition: Visual sentiment
  analysis with deep coupled adjective and noun neural networks, in: IJCAI,
  2016.

\bibitem{Cimpoi2015}
M.~Cimpoi, S.~Maji, A.~Vedaldi, Deep filter banks for texture recognition and
  segmentation, in: IEEE Conference on Computer Vision and Pattern Recognition,
  2015.

\bibitem{Subramanian2011}
R.~Subramanian, V.~Yanulevskaya, N.~Sebe, Can computers learn from humans to
  see better?: inferring scene semantics from viewers' eye movements, in:
  Proceedings of the 19th ACM international conference on Multimedia, MM '11,
  ACM, New York, NY, USA, 2011, pp. 33--42.
\newblock \href {http://dx.doi.org/10.1145/2072298.2072305}
  {\path{doi:10.1145/2072298.2072305}}.

\bibitem{Subramanian2009}
R.~Subramanian, H.~Katti, R.~Huang, T.-S. Chua, M.~Kankanhalli, Automated
  localization of affective objects and actions in images via caption
  text-cum-eye gaze analysis, in: Proceedings of the 17th ACM international
  conference on Multimedia, MM '09, ACM, New York, NY, USA, 2009, pp. 729--732.
\newblock \href {http://dx.doi.org/10.1145/1631272.1631399}
  {\path{doi:10.1145/1631272.1631399}}.

\bibitem{Soleymani2012}
M.~Soleymani, M.~Pantic, T.~Pun, Multimodal emotion recognition in response to
  videos, Affective Computing, IEEE Transactions on 3~(2) (2012) 211--223.
\newblock \href {http://dx.doi.org/10.1109/T-AFFC.2011.37}
  {\path{doi:10.1109/T-AFFC.2011.37}}.

\bibitem{Lowe1999}
D.~Lowe, Object recognition from local scale-invariant features, in: Computer
  Vision, 1999. The Proceedings of the Seventh IEEE International Conference
  on, Vol.~2, 1999, pp. 1150--1157 vol.2.
\newblock \href {http://dx.doi.org/10.1109/ICCV.1999.790410}
  {\path{doi:10.1109/ICCV.1999.790410}}.

\bibitem{R.-Tavakoli2015}
H.~R.-Tavakoli, A.~Atyabi, A.~Rantanen, S.~J. Laukka, S.~Nefti-Meziani,
  J.~Heikkil\"{a}, Predicting the valence of a scene from observers’ eye
  movements, PLoS ONE 10~(9) (2015) e0138198.
\newblock \href {http://dx.doi.org/10.1371/journal.pone.0138198}
  {\path{doi:10.1371/journal.pone.0138198}}.

\bibitem{Bradley1994}
M.~M. Bradley, P.~J. Lang, Measuring emotion: The self-assessment manikin and
  the semantic differential, Journal of Behavior Therapy and Experimental
  Psychiatry 25~(1) (1994) 49 -- 59.

\bibitem{Wadlinger2006}
H.~A. Wadlinger, D.~M. Isaacowitz, Positive mood broadens visual attention to
  positive stimuli, Motivation and Emotion 30.

\bibitem{Tichon2014}
J.~Tichon, T.~M.~G. Wallis, T.~Visser, S.~Riek, Using pupillometry and
  electromyography to track positive and negative affect during flight
  simulation, Aviation Psychology and Applied Human Factors 4~(1).

\bibitem{Simola2015}
J.~Simola, K.~L. Fevre, J.~Torniainen, T.~Baccino, Affective processing in
  natural scene viewing: Valence and arousal interactions in
  eye-fixation-related potentials, NeuroImage 106 (2015) 21 -- 33.
\newblock \href
  {http://dx.doi.org/http://dx.doi.org/10.1016/j.neuroimage.2014.11.030}
  {\path{doi:http://dx.doi.org/10.1016/j.neuroimage.2014.11.030}}.

\bibitem{Lithari2010}
C.~Lithari, C.~Frantzidis, C.~Papadelis, A.~Vivas, M.~Klados,
  C.~Kourtidou-Papadeli, C.~Pappas, A.~Ioannides, P.~Bamidis, Are females more
  responsive to emotional stimuli? a neurophysiological study across arousal
  and valence dimensions, Brain Topography 23~(1) (2010) 27--40.
\newblock \href {http://dx.doi.org/10.1007/s10548-009-0130-5}
  {\path{doi:10.1007/s10548-009-0130-5}}.

\bibitem{Torralba2006}
A.~Torralba, A.~Oliva, M.~Castelhano, J.~Henderson, Contextual guidance of eye
  movements and attention in real-world scenes: the role of global features in
  object search, Psychol Rev 113~(4) (2006) 766--86.

\bibitem{Tatler2007}
B.~Tatler, The central fixation bias in scene viewing: selecting an optimal
  viewing position independently of motor bases and image feature
  distributions., Journal of Vision 14~(7).

\bibitem{Greene2012}
M.~R. Greene, T.~Liu, J.~M. Wolfe, Reconsidering {Y}arbus: A failure to predict
  observers’ task from eye movement patterns, Vision Research 62 (2012) 1 --
  8.
\newblock \href {http://dx.doi.org/10.1016/j.visres.2012.03.019}
  {\path{doi:10.1016/j.visres.2012.03.019}}.

\bibitem{Borji2014a}
A.~Borji, L.~Itti, Defending yarbus: Eye movements reveal observers’ task,
  Journal of Vision 14~(5).

\bibitem{Kanan2014}
C.~Kanan, N.~A. Ray, D.~N.~F. Bseiso, J.~H. Hsiao, G.~W. Cottrell, Predicting
  an observer's task using multi-fixation pattern analysis, in: Proceedings of
  the Symposium on Eye Tracking Research and Applications, ETRA '14, ACM, New
  York, NY, USA, 2014, pp. 287--290.
\newblock \href {http://dx.doi.org/10.1145/2578153.2578208}
  {\path{doi:10.1145/2578153.2578208}}.

\bibitem{Maaten2011}
L.~van~der Maaten, Learning discriminative fisher kernels, in: International
  Conference on Machine Learning (ICML), 2011, pp. 217--224.

\bibitem{LeMeur2011a}
O.~Le~Meur, T.~Baccino, A.~Roumy, {Prediction of the Inter-Observer Visual
  Congruency (IOVC) and Application to Image Ranking}, in: {ACM Multimedia},
  Phoneix, United States, 2011.

\bibitem{Mancas2013}
M.~Mancas, O.~L. Meur, Memorability of natural scenes: The role of attention,
  in: {IEEE} International Conference on Image Processing, {ICIP} 2013,
  Melbourne, Australia, September 15-18, 2013, 2013, pp. 196--200.

\bibitem{Lowe2004}
D.~G. Lowe, Distinctive image features from scale-invariant keypoints, Int. J.
  Comput. Vision 60~(2) (2004) 91--110.
\newblock \href {http://dx.doi.org/10.1023/B:VISI.0000029664.99615.94}
  {\path{doi:10.1023/B:VISI.0000029664.99615.94}}.

\bibitem{Bosch2007}
A.~Bosch, A.~Zisserman, X.~Muoz, Image classification using random forests and
  ferns, in: Computer Vision, 2007. ICCV 2007. IEEE 11th International
  Conference on, 2007, pp. 1--8.
\newblock \href {http://dx.doi.org/10.1109/ICCV.2007.4409066}
  {\path{doi:10.1109/ICCV.2007.4409066}}.

\bibitem{Bay2008}
H.~Bay, A.~Ess, T.~Tuytelaars, L.~V. Gool, Speeded-up robust features ({SURF}),
  Computer Vision and Image Understanding 110~(3) (2008) 346 -- 359, similarity
  Matching in Computer Vision and Multimedia.
\newblock \href
  {http://dx.doi.org/http://dx.doi.org/10.1016/j.cviu.2007.09.014}
  {\path{doi:http://dx.doi.org/10.1016/j.cviu.2007.09.014}}.

\bibitem{Sande2010}
K.~E.~A. van~de Sande, T.~Gevers, C.~G.~M. Snoek, Evaluating color descriptors
  for object and scene recognition, IEEE Transactions on Pattern Analysis and
  Machine Intelligence 32~(9) (2010) 1582--1596.

\bibitem{Simonyan2015}
K.~Simonyan, A.~Zisserman, Very deep convolutional networks for large-scale
  image recognition, in: ICLR, 2015.

\bibitem{Cortes1995}
C.~Cortes, V.~Vapnik, Support-vector networks, Machine Learning 20~(3) (1995)
  273--297.
\newblock \href {http://dx.doi.org/10.1023/A:1022627411411}
  {\path{doi:10.1023/A:1022627411411}}.

\bibitem{He2009}
H.~He, E.~A. Garcia, Learning from imbalanced data, IEEE TRANSACTIONS ON
  KNOWLEDGE AND DATA ENGINEERING 21~(9) (2009) 1263 -- 1284.

\bibitem{Atyabi2012}
A.~Atyabi, M.~Luerssen, S.~Fitzgibbon, D.~M.~W. Powers, Evolutionary feature
  selection and electrode reduction for eeg classification, in: Evolutionary
  Computation (CEC), 2012 IEEE Congress on, 2012, pp. 1--8.
\newblock \href {http://dx.doi.org/10.1109/CEC.2012.6256130}
  {\path{doi:10.1109/CEC.2012.6256130}}.

\bibitem{McNemar1947}
Q.~McNemar, Note on the sampling error of the difference between correlated
  proportions or percentages, Psychometrika 12~(2) (1947) 153 -- 157.

\bibitem{Oliva2001}
A.~Oliva, A.~Torralba, Modeling the shape of the scene: A holistic
  representation of the spatial envelope, International Journal of Computer
  Vision 42 (2001) 145--175.

\bibitem{Siagian2007}
C.~Siagian, L.~Itti, Rapid biologically-inspired scene classification using
  features shared with visual attention, IEEE Transactions on Pattern Analysis
  and Machine Intelligence 29~(2) (2007) 300--312.

\bibitem{Gomez2013}
P.~Gomez, A.~von Gunten, B.~Danuser, Content-specific gender differences in
  emotion ratings from early to late adulthood, Scandinavian Journal of
  Psychology\href {http://dx.doi.org/10.1111/sjop.12075}
  {\path{doi:10.1111/sjop.12075}}.

\bibitem{Judd2012}
T.~Judd, F.~Durand, A.~Torralba, A benchmark of computational models of
  saliency to predict human fixations, Tech. Rep. MIT-CSAIL-TR-2012-001,
  Massachusetts institute of technology (2012).

\bibitem{Deng2009}
J.~Deng, W.~Dong, R.~Socher, L.-J. Li, K.~Li, L.~Fei-Fei, {ImageNet: A
  Large-Scale Hierarchical Image Database}, in: CVPR09, 2009.

\bibitem{Torresani2010}
L.~Torresani, M.~Szummer, A.~Fitzgibbon, Efficient object category recognition
  using classemes, in: Proceedings of the 11th European Conference on Computer
  Vision: Part I, ECCV'10, Springer-Verlag, Berlin, Heidelberg, 2010, pp.
  776--789.

\bibitem{Bradley2007a}
M.~Bradley, P.~J. Lang, The international affective picture system
  ({I}{A}{P}{S}) in the study of emotion and attention, in: J.~A. Coan,
  J.~J.~B. Allen (Eds.), Handbook of Emotion Elicitation and Assessment, Oxford
  University Press, 2007, pp. 29--46.

\bibitem{Zhang2015}
H.~Zhang, M.~Gönen, Z.~Yang, E.~Oja, Understanding emotional impact of images
  using bayesian multiple kernel learning, Neurocomputing 165 (2015) 3 -- 13.
\newblock \href
  {http://dx.doi.org/http://dx.doi.org/10.1016/j.neucom.2014.10.093}
  {\path{doi:http://dx.doi.org/10.1016/j.neucom.2014.10.093}}.

\bibitem{Haji-Abolhassani2014}
A.~Haji-Abolhassani, J.~J. Clark, An inverse yarbus process: Predicting
  observers’ task from eye movement patterns, Vision Research 103 (2014) 127
  -- 142.
\newblock \href
  {http://dx.doi.org/http://dx.doi.org/10.1016/j.visres.2014.08.014}
  {\path{doi:http://dx.doi.org/10.1016/j.visres.2014.08.014}}.

\end{thebibliography}

\end{document}